%% file: tpds_FRuDA.tex
\documentclass[10pt,journal,compsoc]{IEEEtran}

\usepackage[algo2e,vlined,ruled,linesnumbered]{algorithm2e}

\usepackage{wrapfig}
\usepackage{mwe}
\usepackage{textcomp} 
\usepackage{subcaption}
\usepackage{enumitem}
\usepackage{booktabs}
\usepackage{multirow}
\usepackage{xcolor}
\usepackage{bbm}
\usepackage{url}

\usepackage{tabularx}
\usepackage[export]{adjustbox}
\usepackage[font=small,labelfont=bf,tableposition=top]{caption}
\usepackage{diagbox, colortbl}
\usepackage{changepage}
\usepackage{etoolbox}
\usepackage{amssymb}
\usepackage{listings}
\usepackage{amsmath}

\newcommand{\blue}[1]{\textcolor{black}{#1}}

\ifCLASSOPTIONcompsoc
  \usepackage[nocompress]{cite}
\else
  \usepackage{cite}
\fi

\makeatletter
\patchcmd{\@algocf@start}
  {-1.5em}
  {0pt}
  {}{}
\makeatother

\newtheorem{Theorem}{Theorem}

\newsavebox{\bigleftbox}
\input{math_commands.tex}

\newcommand{\parjump}{\vspace{+0.5em}}

\newcommand{\system}{FRuDA}

\let\oldincludegraphics\includegraphics
\newcommand{\scalefiginput}[2][1]{%
  \renewcommand{\includegraphics}[2][]{\scalebox{#1}{\oldincludegraphics[##1]{##2}}}%
  \input{#2}
  \let\includegraphics\oldincludegraphics
}

\begin{document}

\title{\system{}: Framework for Distributed Adversarial Domain Adaptation}

\newcommand{\squishlist}{
 \begin{list}{$\bullet$}
  { \setlength{\itemsep}{0pt}
     \setlength{\parsep}{3pt}
     \setlength{\topsep}{3pt}
     \setlength{\partopsep}{0pt}
     \setlength{\leftmargin}{1em}
     \setlength{\labelwidth}{1em}
     \setlength{\labelsep}{0.5em} } }
\newcommand{\squishend}{\end{list}}

\author{Shaoduo Gan*, Akhil Mathur*, Anton Isopoussu, Fahim Kawsar, Nadia Berthouze, Nicholas D. Lane
\IEEEcompsocitemizethanks{\IEEEcompsocthanksitem Shaoduo Gan is with ETH Zurich and did this work while on an internship with Nokia Bell Labs, Cambridge, UK.
\IEEEcompsocthanksitem Akhil Mathur and Fahim Kawsar are with Nokia Bell Labs, Cambridge, UK.
\IEEEcompsocthanksitem Anton Isopoussu is with Invenia Labs, Cambridge, UK.
\IEEEcompsocthanksitem Nadia Berthouze is with University College London, London, UK.
\IEEEcompsocthanksitem Nicholas D. Lane is with University of Cambridge, Cambridge, UK.
}
\thanks{* denotes joint primary authors}
}


\IEEEtitleabstractindextext{
\begin{abstract}
Breakthroughs in unsupervised domain adaptation (uDA) can help in adapting models from a label-rich source domain to unlabeled target domains. Despite these advancements, there is a lack of research on how uDA algorithms, particularly those based on adversarial learning, can work in distributed settings. In real-world applications, target domains are often distributed across thousands of devices, and existing adversarial uDA algorithms -- which are centralized in nature -- cannot be applied in these settings. To solve this important problem, we introduce \system{}: an end-to-end framework for distributed adversarial uDA. Through a careful analysis of the uDA literature, we identify the design goals for a distributed uDA system and propose two novel algorithms to increase adaptation accuracy and training efficiency of adversarial uDA in distributed settings. Our evaluation of \system{} with five image and speech datasets show that it can boost target domain accuracy by up to 50\% and improve the training efficiency of adversarial uDA by at least {$11\times$}.
\end{abstract}

}

\maketitle
\IEEEdisplaynontitleabstractindextext
\IEEEpeerreviewmaketitle

\IEEEraisesectionheading{\section{Introduction}\label{sec:intro}}
\input{chapters/intro}

\section{Preliminaries and Design Goals} 
\label{sec:problem}
\input{chapters/preliminaries_motivation.tex}

\input{chapters/rw}

\section{\system{}: Framework for Distributed Adversarial Domain Adaptation}
\label{sec:approach}
\input{chapters/approach.tex}


\section{Evaluation}
\label{sec:eval}
\input{chapters/eval.tex}


\section{Conclusion}
\input{chapters/conclusion.tex}


\small
\bibliographystyle{IEEEtran}
\bibliography{tpds}

\clearpage

\appendices
\input{chapters/appendix.tex}

\end{document}

%% file: math_commands.tex
\usepackage{amsmath,amsfonts,bm}

\def\eqref#1{equation~\ref{#1}}

\def\1{\bm{1}}

\DeclareMathAlphabet{\mathsfit}{\encodingdefault}{\sfdefault}{m}{sl}
\SetMathAlphabet{\mathsfit}{bold}{\encodingdefault}{\sfdefault}{bx}{n}



%% file: chapters/intro.tex
Unsupervised Domain Adaptation \textbf{(uDA)} is a sub-field of machine learning aimed at adapting a model trained on a labeled source domain to a different, but related, unlabeled target domain. Over the last few years, \emph{adversarial domain adaptation} has emerged as a prominent method for uDA, wherein the core idea is to learn domain-invariant feature representations from the data using adversarial learning \cite{long2015learning, long2017deep, long2018conditional, ganin2016domain, tzeng2017adversarial, hoffman2018cycada, shen2018wasserstein, zou2019consensus}. 
This paper focuses on exploring adversarial domain adaptation in a \emph{distributed} setting. Prior works~\cite{ganin2016domain, tzeng2017adversarial, hoffman2018cycada, shen2018wasserstein} have taken an algorithmic viewpoint to adversarial domain adaptation and assumed that datasets from the source and target domains are available on the same machine and can be accessed freely during the adaptation process. While this assumption has made it easy to research uDA algorithms, it can be easily violated in real-world scenarios where the domain datasets reside on massively distributed devices and are not allowed to be shared for privacy reasons.

As an example, let us consider the task of developing personalized health monitoring solutions on mobile devices, such as the prediction of COVID-19 from cough sounds collected from a smartphone~\cite{brown2020exploring}. A company (\emph{i.e., a source domain}) can collect a labeled dataset for the task and train a prediction model on it. Later, this source model needs to be deployed for smartphone users around the world (\emph{i.e., target domains}) and will require adapting to each user's personal health condition and biomarkers. As collecting labels from the target domains is challenging in this setting, unsupervised domain adaptation could become a promising approach to adapt the source model and tailor it to each user's data. 

However, since the health data records from target users are distributed across thousands of devices and cannot be uploaded on a central server due to potential privacy reasons, we cannot use the centralized uDA techniques proposed in the literature. Performing adversarial domain adaptation in such distributed settings remains an under-explored problem and is the main focus of this paper. Clearly, if adversarial uDA can be extended to such distributed settings, it will further enhance the potential for real-world impact of these algorithms.

\parjump{}
\noindent
\textbf{Setup and Challenges.} We consider a distributed system where each domain dataset resides on a different node of the system. We assume that one domain dataset is labeled and represents the \emph{source domain}. Other domain datasets are unlabeled and represent the \emph{target domains} which need to undergo domain adaptation to learn a model tailored to their data distribution. New target domains can join the distributed system at any time. The nodes are able to communicate with each other over a network. Based on this novel but realistic setup, we highlight three key challenges in building a distributed adversarial uDA system:

\begin{itemize}[align=left, leftmargin=*]

    \item The first challenge relates to \emph{system design}: how do we design distributed adversarial neural network architectures that can perform uDA without exchanging any raw data between domains. Distributed training techniques have been studied extensively for supervised learning~\cite{lian2017can, tang2018d}, but their investigation for adversarial uDA remains under-explored.
    
    \item Next is the challenge of \emph{efficiency}. In distributed machine learning, any communication between the nodes incurs a cost, both in terms of training time and data transfer expenses. As such, the design of a distributed adversarial uDA system should be as efficient as possible.
    
    \item Finally, a key challenge is to obtain the highest possible \emph{accuracy} for each target domain after adaptation. Let us define a \emph{collaborator} as the domain with which a target domain undergoes adaptation, e.g., in X$\xrightarrow{}$Y adaptation, X is the collaborator for the target domain Y. Seminal theoretical works~\cite{ben2010theory} in domain adaptation have proven that the accuracy obtained for a given target domain is highly dependent on the characteristics of the collaborator with which the adaptation is performed. Hence, it becomes critical to select the \emph{optimal collaborator} for each target domain in a distributed setting, to ensure that the target domain can achieve the highest accuracy through adaptation. 
    
\end{itemize}

\begin{figure}[t]
    \centering
    \begin{subfigure}{0.5\textwidth}%
        \centering
        \includegraphics[scale=0.12]{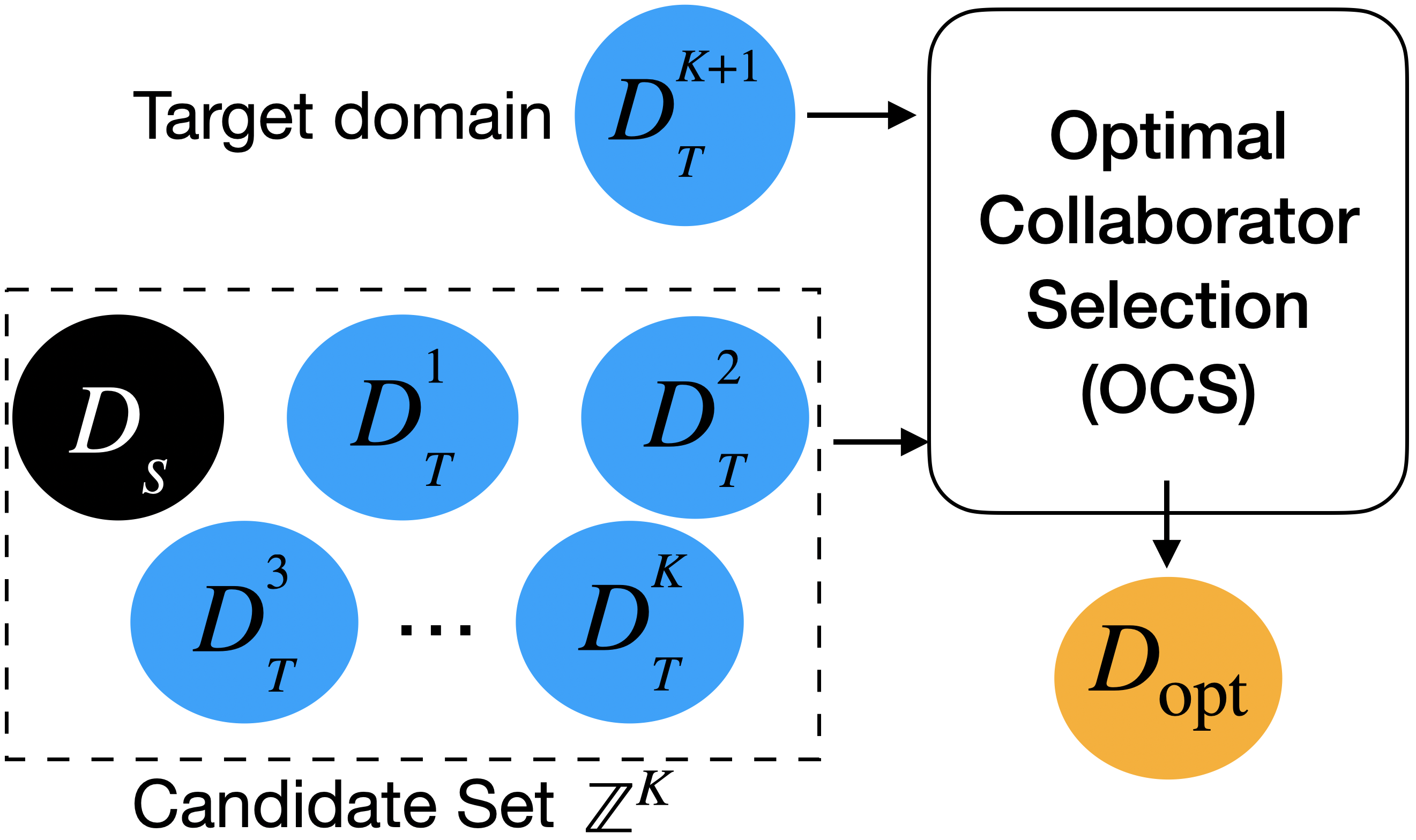}
        \caption{Optimal Collaborator Selection (OCS)}
        \label{fig:ocs_diagram}
    \end{subfigure}%
    \vspace{0.2cm}
    \begin{subfigure}{0.5\textwidth}%
        \centering
        \includegraphics[scale=0.12]{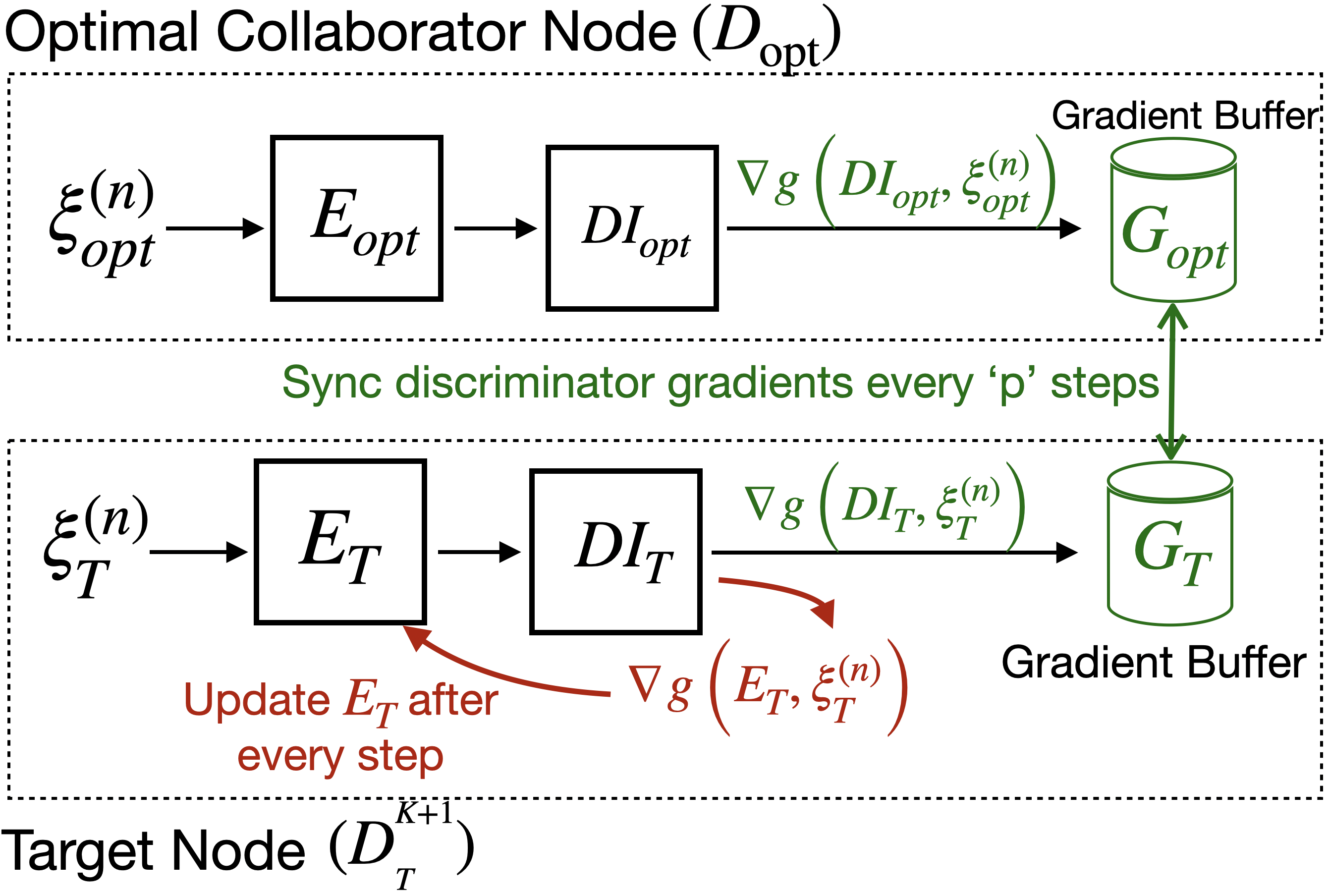}
        \caption{Distributed uDA using DILS}
        \label{fig:dils_diagram}
    \end{subfigure}%
      \vspace{-0.2cm}
\captionsetup{font=small,labelfont=small}
\caption{Illustration of \system{}. (a) A new target domain $D_T^{K+1}$ finds its optimal adaptation collaborator  $D_\mathrm{opt}$ from a set of candidate domains. (b) $D_T^{K+1}$ performs distributed uDA with $D_\mathrm{opt}$ to learn a model for its distribution.}
  \label{fig:fruda_diagram}
  \vspace{-0.3cm}
\end{figure}

\parjump{}
\noindent
\textbf{Contributions.} In this paper, we propose an end-to-end learning framework, named \textbf{F}ramework for \textbf{R}ealistic uDA (FRuDA) which addresses all the above challenges in a unified manner, and makes adversarial uDA accurate, efficient, and privacy-preserving in a distributed setting. Figure \ref{fig:fruda_diagram} demonstrates the system overview of FRuDA. The source code of FRuDA can be found in Appendix A.1. At the core of FruDA are two novel contributions: 

\begin{enumerate}[align=left, leftmargin=*]

\item A distributed collaborator selection algorithm called Optimal Collaboration Selection (OCS) which finds the best adaptation \emph{collaborator} for each unlabeled target domain in the system. OCS is built on a novel theoretical formulation, which selects the optimal collaborator based on the collaborator's own in-domain error and the Wasserstein distance between the collaborator and target domain. Our results show that OCS can lead to an increase in the target domain accuracy in distributed uDA systems by as much as 50\% over various baselines. 

\item A distributed training strategy called Discriminator-based Lazy Synchronization (DILS) which decomposes the adversarial learning architecture across distributed nodes and performs uDA by exchanging the gradients of the domain discriminator between nodes. DILS ensures that unlabeled target domains can learn a prediction model through adaptation from other domains in a privacy-preserving manner, without revealing or exchanging their raw data. DILS also allows for setting a trade-off between the \emph{accuracy} and \emph{efficiency} objectives of distributed uDA, using a tunable parameter.      

\end{enumerate}

The paper is structured as follows. In \S\ref{sec:problem}, we provide a brief primer on adversarial domain adaptation and further motivate the design requirements of a distributed uDA system. In \S\ref{sec:technique}, we present our end-to-end learning framework FruDA, describe the two core algorithms on which FruDA is built, and provide theoretical justifications for our design. In \S\ref{sec:eval}, we provide a comprehensive evaluation of FruDA on multiple vision and speech datasets. We compare FruDA against various baselines for selecting adaptation collaborators and observe that it significantly outperforms them on target domain accuracy, by as much as 50\%. {We also illustrate that FruDA can reduce the amount of data exchanged during training by at least $11\times$ when compared to state-of-the-art baselines}, without significantly compromising the adaptation accuracy. Finally, we show that FruDA can co-exist with various types of adversarial uDA algorithms proposed in the literature, thus making it a generalizable framework for supporting distributed adversarial uDA.

%% file: chapters/preliminaries_motivation.tex
\subsection{Primer on Adversarial uDA}
\label{sec:primer}
 We first provide a brief primer on adversarial uDA with one source and one target domain. In the subsequent sections, we will explain how we extend adversarial uDA to distributed systems with multiple target domains. 
 
 Let $D_S$ be a source domain with labeled training set $\{X_S, Y_S\}$ and $D_T$ be a target domain with unlabeled training set $\{X_T\}$. We can train a feature extractor, $E_S$, and a classifier, $C_S$ for the source domain using supervised learning by optimizing a classification loss $\mathcal R_{cls}$ as follows:
  \begin{align*}
\mathcal R_{cls} = \mathbb{E}_{(x_s, y_s) \sim (X_S, Y_S)} l_{cls}[C_S(E_S(x_s)), y_s)]
\end{align*}, 
where  $l_{cls}(\cdot)$ is a loss function such as categorical cross-entropy. 

The {goal of} adversarial uDA is to learn a feature extractor $E_T$ for the unlabeled target domain, which minimizes the divergence between the empirical source and target feature distributions. If the divergence in feature representations between domains is minimized, we can apply the pre-trained source classifier $C_S$ on the target features and obtain inferences, without requiring to learn a separate target classifier $C_T$. To learn $E_T$, two losses are optimized using adversarial learning, namely the Discriminator Loss $\mathcal R_{dis}$ and the Mapping Loss $\mathcal R_{map}$, as follows:
\begin{align}
\label{eq:gen1}
\mathcal R_{dis} = \mathbb{E}_{x_t \sim X_T, x_s \sim X_S} l_{dis}[DI(E_S(x_s)), DI(E_T(x_t))]
\end{align}

\begin{align}
\label{eq:gen2}
\mathcal R_{map} = \mathbb{E}_{x_t \sim X_T, x_s \sim X_S} l_{map}[DI(E_S(x_s)), DI(E_T(x_t))]
\end{align}

Here $DI$ represents a domain discriminator tasked with separating data from source and target domains.$l_{dis}(\cdot)$ and $l_{map}(\cdot)$ are the adversarial loss functions, which have been studied by several previous works. For example, DANN \cite{ganin2016domain} uses a cross-entropy loss to compute $l_{dis}$, where the labels indicate the domain of the data. For the Mapping loss, they simply compute $l_{map}(\cdot) = -l_{dis}(\cdot)$ using a Gradient Reversal Layer (GRL). Instead, ADDA \cite{tzeng2017adversarial} computes the mapping loss using the label inversion trick. Other works such as \cite{shen2018wasserstein} use Wasserstein distance as the metric to compute $l_{dis}$. 

The important takeaway here is that even though different algorithms employ different loss functions, the general training paradigm of adversarial uDA remains similar as shown in Eq.~\ref{eq:gen1} and \ref{eq:gen2}. This simple yet important insight means that it is possible to design a generalized domain adaptation framework for a distributed setting that can work for many uDA algorithms. This intuition will be confirmed in \S\ref{sec:eval} where we show that our proposed framework can work with four types of uDA optimization objectives. 

\subsection{Design Requirements for a Distributed Domain Adaptation System}
\label{subsec.requirements}
In this section, we describe two key design requirements for a distributed domain adaptation system, which will serve as a guide for the algorithms and training strategies proposed in the paper. 

\parjump{}
\noindent
{\textbf{Training Efficiency.}} Efficiency is a key design metric for any distributed system. Compared with the powerful computation capabilities of modern hardware, communication tends to be the main bottleneck in distributed training. Especially, in distributed domain adaptation, a domain dataset could possibly reside on a smartphone, a laptop or a wearable device, whose communication capabilities are far inferior than data center machines. In such massively distributed scenarios, an optimal communication strategy is crucial for system's efficiency and user experience.

Recall the model components $E_S$, $E_T$, $C_S$, and $DI$ mentioned in \S\ref{sec:primer} that are involved in adversarial uDA. In a non-distributed setting, these components reside on the same machine and passing data from one to another has negligible cost. However, in a distributed setup, each domain has to keep a copy of these components on their machine and any exchange of information between them takes time and incurs a communication cost. 

Since raw data is private and cannot be be exchanged in our problem setup, we exchange the gradients of these model components to facilitate distributed training. Hence, the key challenge is: \textit{what is the most communication-efficient strategy to exchange model gradients across domains?} More specifically, this question can be decomposed into three sub-questions: 1) \textit{how many domains should be involved in the communication?} 2) \textit{which model components are necessary to be communicated?} and 3) \textit{how frequently do they need to be communicated?}

To answer these questions, we analyze various types of adversarial uDA approaches in the literature. Based on the number of domains involved in training, adversarial uDA algorithms can be categorized into pairwise adaptation \cite{ganin2016domain, shen2018wasserstein, zou2019consensus, tzeng2017adversarial}, multi-source adaptation \cite{peng2019federated, zhao2018adversarial}, and multi-target adaptation \cite{ragab2020adversarial}. Clearly, as the latter two approaches require either multiple source or target domains, they will incur higher communication costs and are less efficient than pairwise adaptation. Another way to classify adversarial uDA algorithms is tied \cite{ganin2016domain} or untied \cite{tzeng2017adversarial} algorithms, depending on whether the feature extractors of source and target domains share weights (i.e., tied) or not (i.e., untied). Untied algorithms are communication-efficient because the feature extractors could be trained separately and their gradients do not need to be exchanged during training. Only the gradients of domain discriminator $DI$ need to be shared across nodes. For these reasons, the design of our proposed framework will be based on the \textbf{pairwise and untied} adversarial uDA algorithms. 

After narrowing down the scope of uDA algorithms from an efficiency perspective, we need to decide how frequently should we communicate gradients between nodes. Naively, we can exchange gradients of the discriminator between nodes after training \emph{each} batch of data, which is common in data-parallel distributed training. However, this approach comes at the expense of a significant communication cost. Instead, we will propose a  \textbf{lazy synchronization} strategy that exchanges the gradients every $p$ steps, and further reduces the communication costs of adversarial uDA with negligible impact on adaptation accuracy. Our proposed training strategy is presented in detail in \S\ref{subsec.dils}.

FADA \cite{peng2019federated} is a recently proposed technique for performing adversarial uDA in a federated learning setting. Based on the above analysis, FADA can be characterized as a multi-source uDA approach which exchanges the gradients of feature extractors after every training step. Although FADA originally solves a different problem than ours and assumes that the system has multiple labeled domains, we implement a special case of FADA in \S\ref{sec:eval} to provide a fair comparison of its efficiency with our technique.

\begin{figure}[t]
\centering
\begin{subfigure}{.5\textwidth}
  \centering
  \includegraphics[scale=0.11]{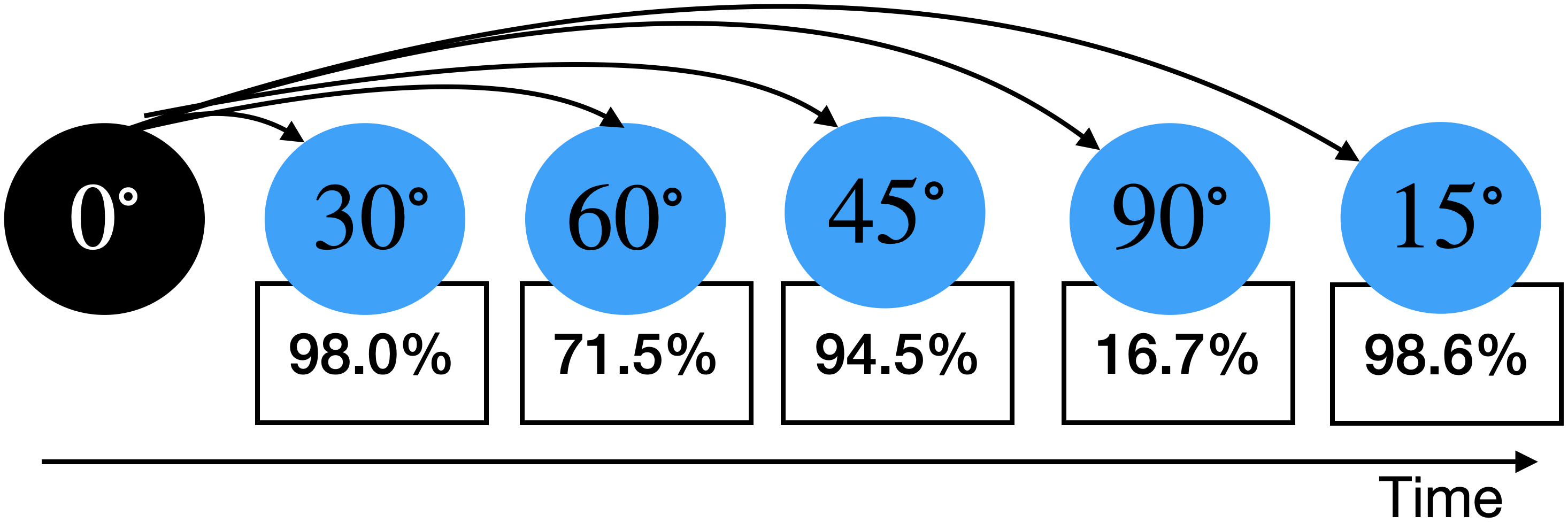}
    \caption{}
  \label{fig:static}
\end{subfigure}%

\begin{subfigure}{.5\textwidth}
  \centering
  \includegraphics[scale=0.12]{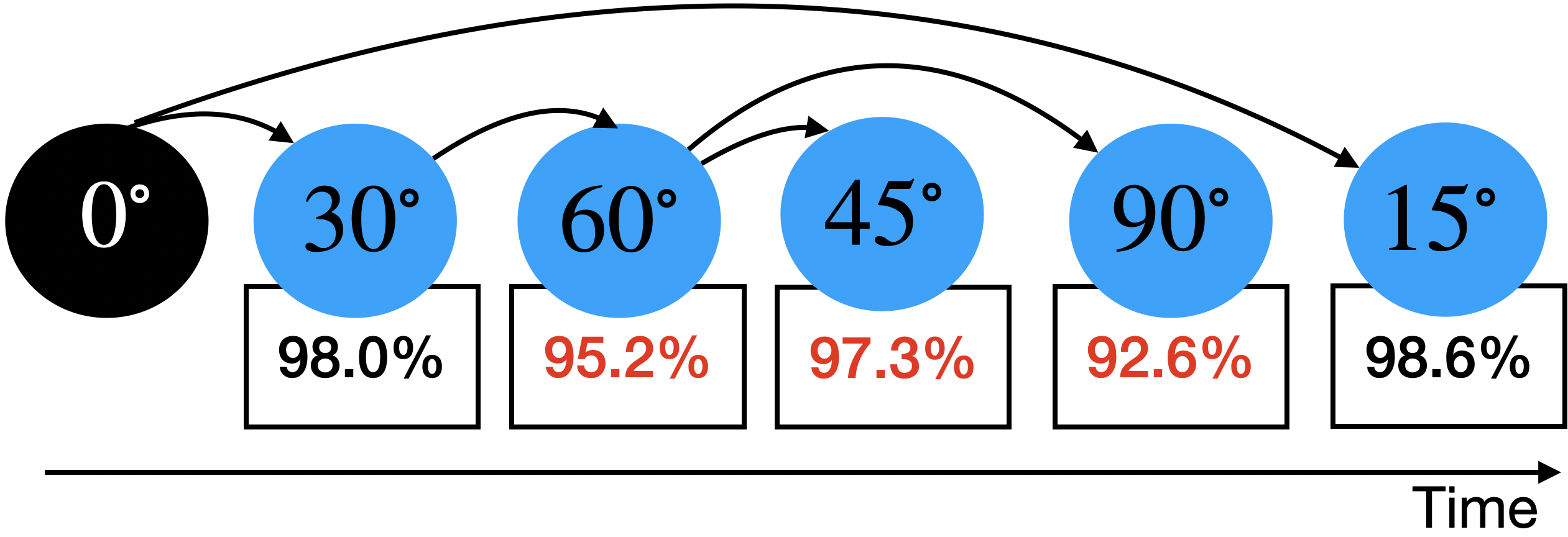}
  \caption{}
  \label{fig:dynamic}
\end{subfigure}
\vspace{-0.3cm}
\caption{0\textdegree{} is the labeled source domain while the domains in blue are unlabeled target domains appearing sequentially in the system. The numbers in rectangle denote the post-adaptation accuracy for a domain. (a) Static Design: Labeled Source acts the collaborator for each target domain. (b) Flexible Design: Each target domain chooses its collaborator dynamically. Previously adapted target domains can also act as collaborators. Note that choosing the right collaborator leads to major accuracy gains over the Static Design for many domains (shown in red).}
\vspace{-0.4cm}
\label{fig:motivation}
\end{figure}

\parjump{}
\noindent
{\textbf{Target Domain Accuracy.}} Obtaining a high accuracy in the target domain is arguably the most important metric of success for a uDA solution. Prior works in learning theory~\cite{ben2010theory} have shown that the classification error obtained in a target domain depends on the characteristics of the collaborator with which the adaptation is performed. This means that if we select the right (or optimal) collaborator for each target domain, it could significantly reduce the target domain error and boost the target domain accuracy. 

To better explain this aspect, we show an experiment on Rotated MNIST, a variant of MNIST in which digits are rotated clockwise by different degrees. Assume that 0\textdegree, i.e., no rotation, is the labeled source domain while 30\textdegree{}, 60\textdegree{}, 45\textdegree{}, 90\textdegree{} and 15\textdegree{} are five unlabeled target domains appearing sequentially, for which we would like to learn a model using adversarial uDA. The na\"ive approach adopted by many existing uDA approaches is to always choose the labeled source domain as the collaborator for a target domain. Figure~\ref{fig:static} shows the performance of such existing approaches if each target domain always chooses the labeled source 0\textdegree{} as its collaborator. While this approach results in high accuracies for 15\textdegree{} and 30\textdegree{}, it performs poorly for other domains. Can we do better? \emph{What if a target domain can adapt not just from the labeled source, but also from other target domains which themselves have undergone adaptation in the past}. Figure~\ref{fig:dynamic} shows that if target domains could flexibly choose their collaborators, they can achieve significantly higher accuracies. E.g., if 90\textdegree{} adapts from 60\textdegree{} (which itself underwent adaptation in the past), it could be achieve an accuracy of 92.6\%, which is almost 75\% higher than what could be achieved by adapting from 0\textdegree{}, as shown in Figure~\ref{fig:motivation}.  

In summary, this example provides a clear insight that we can significantly increase the target domain accuracy if we adapt from the right (or the optimal) collaborator. \textbf{How do we choose such an optimal collaborator?} In some aspects, this problem is similar to the source selection problem~\cite{tan2015transitive} where a metric such as A-distance is employed to compute a distance measure between domains, and the domain with the least distance from the target is chosen as its collaborator. However, none of the prior works are designed for a distributed setup and they require access to raw data samples from both domains to compute the distance metric for source selection. Instead, we propose a fully-distributed approach which selects an optimal collaborator based on the collaborator's in-domain error and the Wasserstein distance between the collaborator and target.

%% file: chapters/rw.tex
\section{Related Work} 
\label{sec:rw}

\textbf{Related work for OCS.} There are prior works on computing similarity between domains, e.g., using distance measures such as Maximum Mean Discrepancy \cite{wang2020transfer}, Gromov-Wasserstein Discrepancy \cite{yan2018semi}, A-distance \cite{wang2018deep}, and subspace mapping \cite{gong2012geodesic}. However, our results in \S\ref{sec:eval} show that merely choosing the most similar domain as the collaborator is not optimal. Instead, OCS directly estimates the target cross-entropy error for collaborator selection. Another advantage of OCS over prior methods is that it relies on Wasserstein distance which could be computed in a distributed setting without exchanging raw data between domains. 

There are also works on selecting or generating intermediate domains for uDA. \cite{tan2015transitive} studied a setting when source and target domains are too distant (e.g., image and text) which makes direct knowledge transfer infeasible. As a solution, they propose selecting intermediate domains using A-distance and domain complexity. However, as we discussed, merely using distance metrics does not guarantee the most optimal collaborator. Moreover, this work was done on a KNN classifier and did not involve adversarial uDA algorithms. \cite{gong2019dlow} and \cite{choi2020visual} use style transfer to generate images in intermediate domains between the source and target. Although interesting, these works are orthogonal to OCS in which the goal is to select the best domain from a given set of candidates. Moreover, these works are primarily focused on visual adaptation, while OCS is a general method that can work for any modality. Finally, \cite{wulfmeier2018incremental, bobu2018adapting} are techniques for incremental uDA in continuously shifting domains. However, in our problem, different target domains may not have any inherent continuity in them and can appear in random order, and hence it becomes important to perform OCS.

\textbf{Related work for DILS.} There is prior work on \emph{distributed model training}, wherein training data is partitioned across multiple nodes to accelerate training. These methods include centralized aggregation \cite{li2014scaling, sergeev2018horovod}, decentralized training \cite{lian2017can, tang2018d}, and asynchronous training \cite{lian2017asynchronous}. Similarly, with the goal of preserving data privacy, \emph{Federated Learning} proposes sharing model parameters between distributed nodes instead of the raw data~\cite{konevcny2016federated, yang2019federated}. However, these distributed and federated training techniques are primarily designed for supervised learning and do not \emph{extend directly to adversarial uDA architectures}. A notable exception is FADA by \cite{peng2019federated} which extends uDA to federated learning. As we discussed earlier, {FADA is designed for a multi-source setting and assumes that all source domains are labeled. Moreover, it exchanges features and gradients of the \emph{feature extractor} between nodes after every step. Instead, DILS operates by doing pairwise adaptation through lazy synchronization of \emph{discriminator gradients} between nodes, which brings significant efficiency gains over FADA.} 

\textbf{Source-free uDA.} \cite{kundu2020towards} and \cite{liang2020we} are two very promising recent works on source-dataset-free uDA. Although the scope of these works are different from us, we share the same goal of making uDA techniques more practical in real-world settings.

\begin{table*}[h]
\centering
\begin{tabular}{|c|c|c|c|c|c|}
\hline
\backslashbox{Property}{Method} & \textbf{\begin{tabular}[c]{@{}c@{}}ADDA~\cite{tzeng2017adversarial},\\ DANN~\cite{ganin2016domain},\\ CADA~\cite{zou2019consensus} \end{tabular}} & \textbf{\begin{tabular}[c]{@{}c@{}}Multi-Source\\ uDA~\cite{zhao2018adversarial, bascol2019improving, cui2020multi}\end{tabular}} & \textbf{\begin{tabular}[c]{@{}c@{}}Federated\\ uDA~\cite{peng2019federated}\end{tabular}} & \textbf{\begin{tabular}[c]{@{}c@{}}Tan et al.\\ ~\cite{tan2015transitive}\end{tabular}} & \textbf{\begin{tabular}[c]{@{}c@{}}\system{}\\ (Ours)\end{tabular}} \\ \hline
\begin{tabular}[c]{@{}c@{}}Adversarial Domain\\ Adaptation\end{tabular} & \checkmark & \checkmark & \checkmark &  & \checkmark \\ \hline
\begin{tabular}[c]{@{}c@{}}Collaborator Selection\\before Adaptation\end{tabular} & &  &  & \checkmark & \checkmark \\ \hline
\begin{tabular}[c]{@{}c@{}}Supports\\Distributed uDA\end{tabular} &  &  & \checkmark &  & \checkmark \\ \hline
\begin{tabular}[c]{@{}c@{}}Communication\\ Efficient Distributed uDA\end{tabular} &  &  &  &  & \checkmark \\ \hline
\begin{tabular}[c]{@{}c@{}}Framework to support \\ multiple uDA algorithms\end{tabular} &  &  &  &  & \checkmark \\ \hline
\begin{tabular}[c]{@{}c@{}}Number of labeled\\ source domains\end{tabular} & 1 & Multiple & Multiple & 1 & 1 \\ \hline
\end{tabular}
\vspace{0.1cm}
\caption{Overview of the related work in unsupervised domain adaptation and the novelty of \system{} over prior works. Check marks denote the core property of different methods. \system{} is unique in providing a framework to scale multiple adversarial uDA algorithms using optimal collaborator selection and privacy-preserving, communication-efficient distributed training.}
\vspace{-2em}
\label{tab:rw}
\end{table*}

%% file: chapters/approach.tex
\label{sec:technique}
\blue{We first present our two algorithmic contributions on Optimal Collaborator Selection (named OCS) and Discriminator-based Lazy Synchronization (named DILS).} Later in \S\ref{subsec.fruda_overview}, we explain how these two algorithms work in conjunction with each other in an end-to-end framework called \system{}, and enable adversarial uDA algorithms to work in a distributed machine learning system.

\subsection{Optimal Collaborator Selection (OCS)}
\label{subsec.ocs}
\input{chapters/colab_selection.tex}

\subsection{Distributed uDA using DIscriminator-based Lazy Synchronization (DILS)} 
\label{subsec.dils}
\input{chapters/distributed.tex}

\subsection{The End-to-End View of \system{}} 
\label{subsec.fruda_overview}
We now discuss how OCS and DILS work together to address the challenges of distributed adversarial uDA introduced in \S\ref{sec:intro}. As shown in Figure~\ref{fig:ocs_diagram}, a new target domain $D_T^{K+1}$ first performs OCS with all candidate domains in $\mathbb{Z}^K$ to find its optimal collaborator $D_\mathrm{opt}$. This step makes uDA systems more flexible and ensures that each target domain is able to achieve the best possible \textbf{adaptation accuracy} in the given setting. Next, as shown in Figure~\ref{fig:dils_diagram}, $D_T^{K+1}$ and $D_\mathrm{opt}$ use DILS to engage in distributed uDA. This step ensures that private raw data is not exposed during adaptation and yet the target domain is able to learn a model for its distribution in an \textbf{efficient} manner. Finally, the newly adapted target domain $D_T^{K+1}$ (with its model and unlabeled data) is added to the candidate set $\mathbb{Z}$ to serve as a potential collaborator for future domains. 

%% file: chapters/colab_selection.tex
Recall from \S\ref{subsec.requirements} that the goal of collaborator selection is to find the optimal collaborator for each target domain. We first formulate the technical problem and then present our solution. 

\subsubsection{Problem Formulation.} 
In our problem setting shown in Figure~\ref{fig:ocs_diagram}, there is one labeled source domain $D_S$ and multiple unlabeled target domains. Let $\{D_T^{j}\big| j=1,\ldots, K\}$ be the $K$ unlabeled target domains for which we want to learn a prediction model using uDA. We assume that target domains join the distributed system sequentially. 

We define a candidate set $\mathbf{\mathbb{Z}_{\tau}}$ as the set of candidate domains that are available to collaborate with a target domain at step $\tau$. When the system initializes at step $\tau = 0$, only the labeled source domain has a learned model, hence $\mathbb{Z}_{0} = \{D_S\}$. When the first target domain $D_{T}^{^1}$ appears, it can only adapt from $D_S$ and learn a model ${E}_{T}^{^1}$. Having learned the model, $D_{T}^{^1}$ is now added to the candidate set $\mathbf{\mathbb{Z}_{\tau}}$ (along with its unlabeled data) and can act as a collaborator for future domains. 

In general, at step $\tau=K$, $\mathbb{Z}_K = \left\{ D_S \right\} \cup \left\{ D_T^{j}\big| j=1,\ldots, K \right\}$ as shown in Figure~\ref{fig:ocs_diagram}. For a new target domain $D_T^{K+1}$, the goal of OCS is to find an optimal collaborator domain $D_\mathrm{opt} \in \mathbb{Z}_K$, such that:
\begin{equation*}
   D_\mathrm{opt} = \operatornamewithlimits{argmin}_{i=1\dotsc|\mathbb{Z}_{K}|} \Phi (\mathbb{Z}_{K}^{i}, D_T^{K+1})
\end{equation*}
where $\mathbb{Z}_{K}^{i}$ is the $i^{th}$ candidate domain in $\mathbb{Z}_K$ and $\Phi$ is a metric that quantifies the error of collaboration between $\mathbb{Z}_{K}^{i}$ and $D_T^{K+1}$. In other words, OCS aims to select a candidate domain from $\mathbb{Z}_K$ which has the least error of collaboration with the target domain. 

\subsubsection{Solution} 
Our key idea is quite intuitive:  the optimal collaborator $D_\mathrm{opt}$ should be a domain, such that adapting from it will lead to the highest classification accuracy (or equivalently, the lowest classification error) in the target domain. We first introduce some notations and then present the key theoretical insight that underpins OCS. 

\fontdimen16\textfont2=3pt
\fontdimen17\textfont2=3pt

\parjump{}
\noindent
\textbf{Notations.} We use domain to represent a distribution $D$ on input space $\mathcal{X}$ and a labeling function $l: \mathcal{X}\rightarrow [0,1]$. A hypothesis is a function $h: \mathcal{X}\rightarrow [0,1]$. Let $\varepsilon_{M, D}\left(h, l\right)$ denote the error of a hypothesis $h$ w.r.t. $l$ under the distribution $D$, where $M$ is an error metric such as $L_1$ error or cross-entropy error .
Further, a function $f$ is called $\theta$-Lipschitz if it satisfies the inequality
$\|f(x) - f(y)\| \leq \theta \|x - y\|$ for some $\theta \in \mathbb{R_+}$. The smallest such $\theta$ is called the Lipschitz constant of $f$.

\parjump{}
\noindent
\textbf{Theorem 1}. Let $D_1$ and $D_2$ be two domains sharing the same labeling function $l$. Let $\theta_{\mathrm{CE}}$ denote the Lipschitz constant of the cross-entropy loss function in $D_1$. For any two $\theta$-Lipschitz hypotheses $h, h'$, we can derive the following error bound for the cross-entropy (CE) error in $D_2$:
\begin{equation}
    \varepsilon_{\mathrm{CE, D_2}}(h, h') \leq  \theta_{\mathrm{CE}} \left( \varepsilon_{\mathrm{L_1, D_1}}(h, h') + 2 \theta W_1(D_1, D_2) \right)
\label{eq:lips}
\end{equation}
where $W_1 (D_1, D_2)$ denote the first Wasserstein distance between the domains $D_1$ and $D_2$, and $\varepsilon_{\mathrm{L_1, D_1}}$ denotes the $L_1$ error in $D_1$. A full proof is provided in the Appendix.

Theorem 1 has two key properties that make it apt for our problem setting. First, it can be used to directly estimate the cross-entropy (CE) error in a target domain ($D_2$), given a hypothesis (or a classifier) from a collaborator domain ($D_1$). Since target CE error is the key metric of interest in classification tasks, this bound is more useful than the one proposed by  \cite{shen2018wasserstein} which estimates the $L_1$ error in the target domain. Secondly, Theorem 1 depends on the Wasserstein distance metric between the domains, which could be computed in a \textbf{distributed} way without exchanging any private data between domains. This property is very important to our distributed problem setup and differentiates it from other distance metrics such as A-distance or Maximum Mean Discrepancy (MMD) which cannot be computed in a distributed manner.

\parjump{}
\noindent
\textbf{Selecting the optimal collaborator.} Motivated by Theorem 1, we now discuss how to select the optimal collaborator for a target domain. Given a collaborator domain $D_c$, a learned hypothesis $h^{c}$ and a labeling function $l$, we can estimate the CE error for a target domain $D_T$ using Theorem 1 as:
    \begin{align}
	 \varepsilon_{\mathrm{CE, D_T}}\left(h^{c}, l\right) \leq \theta_{\mathrm{CE}} (\varepsilon_{\mathrm{L_1, D_c}}\left(h^{c}, l\right) + 2\theta W_1\left(D_c, D_T\right))
    \label{eq:bound_cs}
    \end{align}

We can tighten this bound to get a more reliable estimate of the target CE error. This is achieved by reducing the Lipschitz constant ($\theta$) of the hypothesis $h^{c}$ during training. In uDA, the hypothesis is parameterized by a neural network, and we can train neural networks with small Lipschitz constants by regularizing the spectral norm of each network layer as implemented in \cite{gouk2018regularisation}. Our empirical results show that the upper bound in Eq.~\ref{eq:bound_cs} is a good approximation of the target domain error for the purpose of collaborator selection. In our implementation, we make a simplifying assumption about the availability of a small test set on which the collaborator error can be computed. In future work, we will study more sophisticated error propagation strategies across target domains.

Now that we have a way to estimate the target CE error, we use it to select an optimal collaborator that yields the minimum target CE error. Let $\mathbb{Z} = \{D^k | k=1,\dots,K\}$ be a set of candidate domains each with a pre-trained model $h^k$ with Lipschitz constants $\theta_{\mathrm{CE}}^{k}$ and $\theta^{k}$. Let $D_T^{K+1}$ be a target domain for which the collaborator is to be chosen. We use Eq. \ref{eq:bound_cs} to select the optimal collaborator $D_\mathrm{opt}$:
\begin{equation}
\label{eq:cs_final}
   D_\mathrm{opt} = \operatornamewithlimits{argmin}_{k=1,\dotsc,K} \theta_{\mathrm{CE}}^k( \varepsilon_{\mathrm{L_1, D^{k}}}(h^{k}, l) + 2\theta^{k} W_1(D^{k}, D_{T}^{\mathrm{K+1}}))
\end{equation}

\subsubsection{Computing Wasserstein Distance across Distributed Datasets}
\label{subsec.dist_wass}
Our optimal collaborator selection algorithm requires computing an estimate of the Wasserstein ($W_1$) distance between a candidate domain ($D_C$) and the target domain ($D_T$). Let  $X_{C}$ and  $X_{T}$ denote the unlabeled datasets from the two domains. As shown by \cite{shen2018wasserstein}, the $W_1$ distance can be computed as:  
\begin{equation}
\begin{aligned}
\text{$W_1$} (X_{C}, X_{T}) = &\frac{1}{n_{C}} \sum_{x_{s} \sim \mathcal X_{C}} DI(E_{C}(x_{c})) - \\ &\frac{1}{n_{T}} \sum_{x_{t} \sim \mathcal X_{T}} DI(E_{T}(x_{t}))
\label{eq:wass_impl}
\end{aligned}
\end{equation}

where $n_{C}$ and $n_{T}$ are the number of samples in the dataset, $E_{C}$ and $E_{T}$ are the feature encoders of each domain, and $DI$ is an optimal discriminator trained to distinguish the features from the two domains. To train the optimal discriminator, following loss is minimized: 
\begin{equation*}
\min\limits_{DI} \mathcal L_{adv_{DI}} = -{\mathbb{E}}_{x_{c} \sim \mathcal X_{C}, x_{t} \sim \mathcal X_{T}} [W_1 (x_{c}, x_{t}) + \gamma L_{grad}]
\end{equation*}
where $L_{grad}$ is the gradient penalty used to enforce 1-Lipschitz continuity on the discriminator. 

Interestingly, Equation~\ref{eq:wass_impl} has a similar structure to the optimization objectives for ADDA and other uDA algorithms discussed in the paper. Hence, we can use the same principle as DILS (described in \ref{subsec.dils}) and exchange discriminator gradients between nodes to compute the Wasserstein Distance in a distributed manner, without requiring any exchange of raw data. 

We initialize $E_T$ with $E_C$ and decompose the discriminator $DI$ into two parts ($DI_C$ and $DI_T$) which reside on the respective nodes. The raw data from both nodes is fed into their respective encoders and discriminators, and we compute the gradients of each discriminator as follows:

\begin{equation*}
\mathcal L_{DI}^{^C}  = \frac{1}{n_{C}} \sum_{x_{s} \sim \mathcal X_{C}} DI_C(E_{C}(x_{c})) 
\end{equation*}
\begin{equation*}
\mathcal L_{DI}^{^T} = \frac{1}{n_{T}} \sum_{x_{t} \sim \mathcal X_{T}} DI_T(E_{T}(x_{t}))
\end{equation*}

\begin{minipage}{.5\linewidth}
\begin{equation*}
\nabla g(DI_C, x_c) = \frac{\delta \mathcal L_{DI}^{^C}}{\delta DI_C}
\end{equation*}
\end{minipage}%
\begin{minipage}{.5\linewidth}
\begin{equation*}
\nabla g(DI_T, x_t) = \frac{\delta \mathcal L_{DI}^{^T}}{\delta DI_T}
\end{equation*}
\end{minipage}

Both nodes exchange their discriminator gradients during a synchronization step and compute aggregated gradients:  

\begin{equation}
\nabla g(DI_\mathrm{agg}, x_c, x_t) = \nabla g(DI_C, x_c) - \nabla g(DI_T, x_t)
\end{equation}

Finally, both discriminators $DI_C$ and $DI_T$ are updated with these aggregated gradients, and gradient penalty is applied to enforce the 1-Lipschitz continuity on the discriminators. This process continues until convergence and results in an optimal discriminator. Once the discriminators are trained to convergence, we can calculate the Wasserstein distance as:

\begin{equation*}
\text{$W_1$} (X_{C}, X_{T}) = \mathcal L_{DI}^{^C}  - \mathcal L_{DI}^{^T}
\end{equation*}

%% file: chapters/distributed.tex
\subsubsection{Algorithm description}
Upon selecting an optimal collaborator $D_\mathrm{opt}$ for the target domain $D_T^{K+1}$, the next step is to learn a model for $D_T^{K+1}$ by doing pairwise adversarial uDA with $D_\mathrm{opt}$. In line with our problem setting, both domains are located on distributed nodes and cannot share their training data with each other. 

\input{chapters/algos.tex}

As shown in Figure~\ref{fig:dils_diagram}, we split the adversarial architecture across the distributed nodes. The feature encoders of the collaborator ($E_\mathrm{opt}$) and target ($E_T$) reside on their respective nodes, while the discriminator $DI$ is split into two components $DI_\mathrm{opt}$ and $DI_T$. As we highlighted in \S\ref{sec:problem}, our framework is based on the untied adversarial uDA algorithms and exchanges information between distributed nodes using gradients of the discriminators. Our training process works in two steps: 1) update the target feature extractor $E_T$ to optimize the mapping loss $\mathcal R_{map}$, 2) update the discriminators $DI_T$ and $DI_{opt}$ to optimize the discriminator loss $\mathcal R_{dis}$. Note that $E_{opt}$ is assumed to be pre-trained in the past and is not updated.

We present our DIscriminator-based Lazy Synchronization (DILS) strategy in Algorithm 1. Specifically, at each training step $n$, both nodes feed their domain data $\xi_\mathrm{opt}^{n}$ and $\xi_T^{n}$ into their extractors and discriminators, and compute the gradients of $E_T$, $DI_T$ and $DI_{opt}$, i.e., $\nabla g(E_T, \xi_T^{n})$, $\nabla g(DI_\mathrm{opt}, \xi_\mathrm{opt}^{n})$ and $\nabla g(DI_T, \xi_T^{n})$ respectively. Since $DI_T$ and $DI_{opt}$ are supposed to be shared between nodes, how to synchronize these two components is crucial to adaptation accuracy and communication cost.

If we exchange gradients of $DI_T$ and $DI_{opt}$ after every training step, we can keep the discriminators strictly synchronized and ensure that distributed training converges to the same loss as the non-distributed case. However, this approach has a major downside from an efficiency perspective, as exchanging gradients in every step incurs significant communication costs and increases the overall uDA training time. To boost the training efficiency, we propose a \textit{Lazy Synchronization} approach, wherein instead of every step, the discriminators are synchronized every $p$ training steps, thereby reducing the total communication amount by a factor of $p$. We denote the training steps at which the synchronization takes place as the \emph{sync-up steps} while other steps are called \emph{local steps}. 

In effect, DILS uses synced stale gradients ($\nabla g_{\mathrm{sync}}$) from the latest sync-up step to update the discriminators (line 11), instead of using their local gradients. There are two reasons behind this design choice. Firstly, $DI_T$ and $DI_{opt}$ are two replicas of the same component and are intended to be consistent. Updating them with different local gradients can cause them to diverge. Secondly, the local gradients of $DI_T$ or $DI_{opt}$ are derived from data of only one domain, and are likely to be biased to that domain. Our design choice of applying the latest synced gradients ($\nabla g_{\mathrm{sync}}$) circumvents these limitations of local gradients and guarantees the convergence of adversarial uDA algorithms. Our experiment results also confirm that this approach significantly reduces the uDA communication cost without degrading the target accuracy. 

\subsubsection{Convergence Analysis}
The network structure of \textit{Lazy Synchronization} can be taken as a typical Generative Adversarial Net (GAN) \cite{goodfellow2014generative} where the generative model is target encoder $E_T$ and the discriminative model is discriminator $DI$ ($DI_S$ and $DI_T$). $E_S$ and $E_T$ can separately define two probability distributions $E_S(x_s), x_{s} \sim \mathcal X_{S}$ and $E_T(x_t), x_{t} \sim \mathcal X_{T}$, noted as $p_s$ and $p_t$ respectively. Then according to the theoretical analysis in \cite{goodfellow2014generative}, we know that: 

\textbf{Proposition 1.} For a given $E_T$, if $DI$ is allowed to reach its optimum, and $p_t$ is updated accordingly to optimize the value function, then $p_t$ converges to $p_s$, which is the optimization goal of $E_T$.

In other words, if we can guarantee that under the updating strategy of DILS, convergence behaviours of $DI_S$ and $DI_T$ are close enough to the non-distributed case, then the $E_T$ should be able to converge as well. Note that although $DI_S$ and $DI_T$ are lazily synced, their weights are always identical because they are equally initialized and apply same gradients (latest synced one) for every training step. Therefore, we can take it as one discriminator $DI_{lazy}$ in the convergence analysis. Then we have the following theorem.

\textbf{Theorem 2.} In DILS, given a fixed target encoder, under certain assumptions, we have the following convergence rate for the discriminators $DI_S$ and $DI_T$ (full proof is in Appendix B):

\begin{multline}
\frac{1}{T}\sum_{t=1}^{T}\mathbb{E}[\left \| \nabla f(x_t) \right \|_2^2] \leq \frac{1}{1-\mu L} [ \frac{f(x_0)-f(x_{t+1})}{\mu T}+\\(2p-1)L\mu \sigma ^2+\frac{\mu L\sigma^2}{2}+\frac{\mu^3L^3\sigma^2(2p-1)}{2}]
\end{multline}

Where $p$ is the sync-up step, $\mu$ is the learning rate. Set $\mu = O(1/\sqrt{T})$. When $p << L\sigma^2$, the impact of stale update will be very small, and thus it can converge with rate $O(1/\sqrt{T})$, which is same as the classic SGD algorithm. 

\subsubsection{Privacy Analysis of DILS} 
\label{subsec:dils.privacy}
Recall that a key feature of our distributed training algorithm is to exchange information between the distributed nodes using \emph{gradients of the discriminators}. This clearly affords certain privacy benefits over existing uDA algorithms since we no longer have to transmit raw training data between nodes. However, prior works have shown that model gradients can potentially leak raw training data in collaborative learning \cite{melis2019exploiting}, therefore it is critical to examine: \emph{can the discriminator gradients also indirectly leak training data of a domain}? 

We study the performance of DILS under a state-of-the-art gradient leakage attack proposed by \cite{zhu2019deep}. They showed that gradient matching can be a simple but robust technique to reconstruct the raw training data from stolen gradients. Let us say we are given a machine learning model $F()$ with weights $W$. Let $\nabla W$ be the gradients with respect to a private input pair $(\mathbf{x},y)$. During distributed training, $\nabla W$ are exchanged between the nodes. 

A reconstruction attack happens as follows: an attacker first randomly initializes a dummy input $\mathbf{x'}$ and label input $y'$. This data is fed into the model $F()$ to compute dummy gradients as follows:

$$
\nabla W^{\prime}=\frac{\partial \ell\left(F\left(\mathbf{x}^{\prime}, W\right), \mathbf{y}^{\prime}\right)}{\partial W}
$$

Finally, the attacker minimizes the distance between the dummy gradients and the actual gradients using gradient descent to reconstruct the private data as follows:

\begin{equation}
\begin{aligned}
\mathbf{x}^{\prime *}, \mathbf{y}^{\prime *}&=\underset{\mathbf{x}^{\prime}, \mathbf{y}^{\prime}}{\arg \min }\left\|\nabla W^{\prime}-\nabla W\right\|^{2} \\
&=\underset{\mathbf{x}^{\prime}, \mathbf{y}^{\prime}}{\arg \min }\left\|\frac{\partial \ell\left(F\left(\mathbf{x}^{\prime}, W\right), \mathbf{y}^{\prime}\right)}{\partial W}-\nabla W\right\|^{2}
\label{eq:attack}
\end{aligned}
\end{equation}

\textbf{Can this attack succeed on DILS?} There are two key assumptions in this attack: (i) the weights $W$ of the end-to-end machine learning model are available to an adversary in order for them to compute the dummy gradients, (ii) the gradients of all the layers ($\nabla W$) between the input $x$ and output $y$ are available to the adversary.  

DILS never exchanges the weights of the target domain model (i.e., the feature extractor and the discriminator) during the adversarial training process. As shown in Algorithm 1, the target feature extractor is trained locally and only discriminator gradients are exchanged. Without the knowledge of the model weights $W$, an attacker can not generate the dummy gradients $\nabla W'$ necessary to initiate the attack on the target domain. 

Looking at the source or collaborator domain, we do exchange its feature extractor with the target domain in the initialization step of uDA, which could be used by the attacker to generate the dummy gradients $\nabla W'$. However, for the attack to succeed, the attacker also needs the real gradients ($\nabla W$) of all the layers between the input $x$ and output $y$ in the source domain. This includes gradients of $E_S$ and $DI_S$. In DILS however, we only exchange the gradients of the domain discriminator $DI_S$ during training; the gradients of $E_S$ are never exchanged. Without the knowledge of the gradients of $E_S$, an attacker cannot use Eq.~\ref{eq:attack} to reconstruct the training data of the source domain. 

In summary, we have proven that our strategy of distributed uDA based on discriminator gradients does not allow an attacker to reconstruct the private data of either the source or the target domain.

%% file: chapters/algos.tex
\begin{algorithm2e}[t]
\caption{DILS}
\small
\SetAlCapNameFnt{\small}
\SetAlCapFnt{\Small}
\KwResult{$E_T$}
\textbf{Input}: Pre-trained $E_\mathrm{opt}$; Randomly Initialize $DI_\mathrm{opt}$; Initialize $E_T$ = $E_\mathrm{opt}$; $DI_T$=$DI_\mathrm{opt}$; Sync up step $p$; total steps $N$ \; 
\For{$n = 1,2,...,N$}{
Sample a batch of data on both nodes, $\xi_\mathrm{opt}^{(n)}$ and $\xi_T^{(n)}$. Then feed $\xi_\mathrm{opt}^{(n)}$ and $\xi_T^{(n)}$\ into the respective Extractor-Discriminator model separately on both nodes\;

Based on different loss functions, calculate the gradients locally. On collaborator node, calculate $\nabla g(DI_\mathrm{opt}, \xi_\mathrm{opt}^{(n)})$; On target node, calculate $\nabla g(E_T, \xi_T^{(n)})$ and $\nabla g(DI_T, \xi_T^{(n)})$\;

Add $\nabla g(DI_\mathrm{opt}, \xi_\mathrm{opt}^{(n)})$ to gradient buffer $G_\mathrm{opt}$, add $\nabla g(DI_T, \xi_T^{(n)})$ to target gradients buffer $G_T$ \;
\If{isTargetNode}{
Apply $\nabla g(E_T, \xi_T^{(n)})$ to $E_T$\;
}
\If{$n\%p==0$}{
Exchange gradients buffer and update the latest synced gradients $g_{\mathrm{sync}} = \frac{G_\mathrm{opt}+G_T}{2p}$ \;
Clear $G_\mathrm{opt}$ and $G_T$ \;
}
Apply $g_{\mathrm{sync}}$ to $DI_\mathrm{opt}$ and $DI_T$ separately\;

}
\label{algo:decuda}
\end{algorithm2e}

%% file: chapters/eval.tex
\begin{table*}[t]
\centering
\caption{Mean adaptation accuracy obtained over all target domains in a given order. The sync-up step size $p=4$ is used for this experiment.}
\scalebox{0.85}{
\begin{tabular}[b]{c|c|c||c|c||c|c||c|c||c|c}
\toprule
\multicolumn{1}{l}{} & \multicolumn{2}{c}{\textbf{RMNIST}} & \multicolumn{2}{c}{\textbf{Digits}} & \multicolumn{2}{c}{\textbf{Office-Caltech}} & \multicolumn{2}{c}{\textbf{Mic2Mic}} & \multicolumn{2}{c}{\textbf{DomainNet}} \\
\toprule
 & $Order_{1}$ & $Order_{2}$ & $Order_{1}$ & $Order_{2}$ & $Order_{1}$ & $Order_{2}$ & $Order_{1}$ & $Order_{2}$ & $Order_{1}$ & $Order_{2}$ \\
 \midrule
No Adaptation & 34.65 & 35.54 & 59.59 & 72.09 & 66.40 & 85.25 &  76.45 & 75.83 & 26.12 & 28.32\\
Random & 28.66$\pm$6.50 & 37.11$\pm$4.32  & 62.77$\pm$2.19 & 69.13$\pm$4.0 & 69.18$\pm$1.51 & 80.1$\pm$2.44 & 80.17$\pm$1.60 & 77.34$\pm$1.09 & 20.66$\pm$3.10 & 28.15$\pm$2.89 \\
Labeled Source (LS) & 47.14 $\pm$ 0.85 & 49.08$\pm$ 0.75 & 64.89$\pm$0.23 & 79.87$\pm$0.31 & 67.77$\pm$0.15 & \textbf{90.62$\pm$0.13} & 80.86 $\pm$ 0.09 & 79.91$\pm$0.05 & {33.51$\pm$0.09} & \textbf{36.44$\pm$0.14} \\
Multi-Collaborator & 40.51$\pm$0.30 &  42.73$\pm$0.39 & 60.94$\pm$0.13 & 75.91$\pm$0.30 & 68.90$\pm$0.24 & 82.17$\pm$0.87 & 76.90 $\pm$ 0.13 & 79.0$\pm$0.24 & 18.41$\pm$1.04 & 25.46$\pm$0.61 \\
Proxy A-Distance & 93.51 $\pm$ 0.22 & 74.14$\pm$0.05 & 70.09$\pm$0.45 & 83.07$\pm$0.15 & 69.37$\pm$0.2 & \textbf{90.62$\pm$0.13} & 80.34 $\pm$ 0.19 & 80.02$\pm$0.31 & 35.03$\pm$0.21 & 36.0$\pm$0.49 \\ 
\midrule
\system{} (Ours) & \textbf{97.08 $\pm$ 0.14} & \textbf{81.72$\pm$0.3} & \textbf{73.01$\pm$0.87} & \textbf{85.31$\pm$0.26} & \textbf{74.56$\pm$0.52} & \textbf{90.62$\pm$0.13} & \textbf{81.43 $\pm$ 0.06} & \textbf{81.81$\pm$0.10} & \textbf{37.10$\pm$0.28} & 35.46$\pm$0.31 \\ 
\bottomrule
\end{tabular}
}
\vspace{-0.4cm}
\label{tab:multistep}
\end{table*}

\subsection{Setup}
\textbf{Datasets.} We evaluate \system{} on five image and speech datasets: 

\begin{itemize}[leftmargin=*]
    \item \textbf{Rotated MNIST}: A variant of MNIST with digits rotated clockwise by different degrees. Each rotation is considered a separate domain.
    \item \textbf{Digits}:  Five domains of digits: MNIST (M), USPS (U), SVHN (S), MNIST-M (MM) and SynNumbers (SYN), each consisting of digit classes ranging from 0-9.
    \item \textbf{Office-Caltech}: Images of 10 classes from four domains: Amazon (A), DSLR (D), Webcam (W), and Caltech-256 (C).
    \item \textbf{Mic2Mic}: A speech keyword detection dataset recorded with four microphones: Matrix Creator (C), Matrix Voice (V), ReSpeaker (R) and USB (U). Each microphone represents a domain.
    \item \textbf{DomainNet}: A new challenge dataset from which we use four labeled image domains containing 345 classes each:  Real (R), QuickDraw (Q), Infograph (I), and Sketch (S).
\end{itemize}

\parjump{}
\noindent
\textbf{Evaluation Metrics and Implementation.} \system{} is designed for a distributed system where each node represents a domain. Initially, one labeled source domain $\mathbf{D_S}$ is given in the system and thereafter unlabeled target domains appear sequentially in a random order. The overall accuracy of the system is measured by the mean adaptation accuracy obtained over all target domains. The communication cost is measured by the amount of traffic due to gradient exchange between nodes.
For our experiments, we use a Nvidia V100 GPU to represent a distributed node. All nodes are connected via TCP network. We use Message Passing Interface
(MPI) as the communication primitive between nodes. The system is implemented with TensorFlow 2.0. We use the implementation provided by \cite{gouk2018regularisation} for calculating Lipschitz constants of a neural network. More details on network architectures, hyper-parameters and downloading the source code are provided in the Appendix.

\subsection{Adaptation Accuracy of \system{}}
Let \{$\mathbf{D_S}, D_T^{1}, D_T^{2} \cdots D_T^{K}$\} denote an ordering of one labeled source domain $\mathbf{D_S}$ and $K$ unlabeled target domains. For each target domain $D_T^{i}\lvert_{i=1}^{K}$, we first choose a collaborator domain, which could be either the labeled source domain $\mathbf{D_S}$ or any of the previous target domains $D_T^{j}\lvert_{j=1}^{i-1}$ that have already learned a model using uDA. Upon choosing a collaborator (using OCS or any of the baseline techniques), we use DILS ($p=4$) to perform distributed adversarial uDA between the target domain and the collaborator, and compute the test accuracy $\mathrm{Acc}_T^{i}$ in the target domain. We report the mean adaptation accuracy obtained over all target domains, i.e., $ \frac{1}{K} \sum_{i=1}^K \mathrm{Acc}_T^{i}$. 

For each dataset, we choose two random orderings of source and target domains as shown in Table~\ref{tab:orderings}. The optimization objectives of ADDA~\cite{tzeng2017adversarial} as discussed in \S\ref{sec:primer} are used for adaptation in this experiment.

\textbf{Collaborator Selection Baselines.} We use four baselines for collaborator selection: (i) \emph{Labeled Source} wherein each target domain only adapts from the labeled source domain; (ii) \emph{Random Collaborator}: each target domain chooses a random collaborator from the available candidates; (iii) \emph{Proxy A-distance (PAD)} where we choose the domain which has the least PAD \cite{ben2007analysis} from the target; (iv) \emph{Multi-Collaborator} is based on MDAN \cite{zhao2018adversarial}, where all available candidate domains contribute to the adaptation in a weighted-average way. While this baseline obviously is less efficient than pairwise adaptation, we are interested in comparing its accuracy with our system. Note that MDAN was originally developed assuming that all candidate domains are \emph{labeled}, which is not the case in our setting. Hence, for a fair comparison, we modify MDAN by only optimizing its adversarial loss during adaptation.

\begin{table}[b]
\centering
\caption{Domain orderings used in our experiments. Domains in bold correspond to the labeled source domain, which is introduced first in the system. All other domains have no training labels.}
\scalebox{1.0}{
\begin{tabular}{|c|c|c|}
\hline
\textbf{Dataset} & \textbf{Order 1} & \textbf{Order 2} \\ \hline
RMNIST & \begin{tabular}[c]{@{}c@{}}\textbf{0},30,60,90,120,150,180,\\ 210,240,270,300,330\end{tabular} & \begin{tabular}[c]{@{}c@{}}\textbf{0},180,210,240,270,300,\\ 330,30,60,90,120,150\end{tabular} \\ \hline
Digits & \begin{tabular}[c]{@{}c@{}}\textbf{MM}, Syn, M, U, S \end{tabular} & \begin{tabular}[c]{@{}c@{}}\textbf{Syn}, M, MM, U, S\end{tabular} \\ \hline
Office-Caltech & \textbf{D}, W, C, A & \textbf{W}, C, D, A \\ \hline
Mic2Mic & \textbf{V}, U, C, R & \textbf{U}, C, V, R \\ \hline
DomainNet & \textbf{S}, R, I, Q & \textbf{S}, Q, I, R \\ \hline
\end{tabular}
}
\label{tab:orderings}
\end{table}

\textbf{Results.} Table~\ref{tab:multistep} reports the mean accuracy obtained across all target domains for two orderings in each dataset. \blue{We can observe that collaborator selection techniques have a crucial impact on the adaptation accuracy, and \system{} outperforms the baseline techniques in almost all cases. Below we describe our key findings:}

\blue{Firstly, we observe that \emph{Random Collaborator} and \emph{Multi-Collaborator} are in general the worst among all the methods. We surmise that this is due to the fact that these methods do not consider the distances between domains, and end up with collaborator domains that are too different from the target domain. Compared with \emph{Proxy A-distance (PAD)}, \system{} has better performance because we jointly consider the in-domain error and the Wasserstein distance between domains during collaborator selection.}

\blue{Next, for \emph{Labeled Source} (LS), we observe that its relative performance against \system{} depends on the characteristics of the tasks, and the number and nature of domains. In RMNIST, \system{} provides 41\% accuracy gains over LS on average --- this could be partly attributed to the large number of target domains ($K=11$) in this dataset. As the number of target domains increase, there are more opportunities for benefiting from collaboration selection, which leads to higher accuracy gains over LS in this dataset. For Office-Caltech (Order 2), the performance of \system{} is the same as the Labeled Source baseline because here the the labeled source domain \textbf{W} turned out to be the optimal collaborator for all target domains. Overall, out of the 10 different orderings across 5 datasets, we found that LS outperforms \system{} in only one setting (DomainNet Order 2). For this order, we found that the labeled source: Sketch (S) is the best adaptation collaborator for all subsequent domains, hence the LS performs the best. \system{} makes one error in collaborator selection here: for target domain = Infograph (I), it picks Quickdraw (Q) as the collaborator instead of picking the labeled source Sketch (S), which leads to a 1\% mean accuracy drop.} 

In general, our results demonstrate that as uDA systems scale to multiple target domains, the need for choosing the right adaptation collaborator becomes important, hence warranting the need for accurate collaboration selection algorithms.

\textbf{Generalization to other uDA optimization objectives.} The results presented in Table~\ref{tab:multistep} used the optimization objectives of ADDA~\cite{tzeng2017adversarial}. However, as we discussed earlier, \system{} is intended to be a general framework for distributed uDA and not limited to one specific uDA algorithm. We now evaluate \system{} with three other uDA loss formulations (i) \cite{ganin2016domain}, which uses a Gradient Reversal Layer (GRL) to compute the mapping loss, (ii) \emph{WassDA}\cite{shen2018wasserstein}, which uses Wasserstein Distance as a loss metric for the domain discriminator  and (iii) CADA \cite{zou2019consensus} which operates by enforcing consensus between source and target features. \blue{In Table~\ref{tab:gen}, we observe that different uDA techniques yield different target accuracies after adaptation, depending on their optimization objective. Regardless, \system{} can work in conjunction with all of them to improve the target accuracy over the \emph{Labeled Source} baseline because our proposed framework is designed to be agnostic to the learning algorithm.}

\parjump{}
\noindent
\textbf{Takeaways.} This section highlighted the accuracy gains achieved by \system{} in a distributed uDA setting by selecting the optimal collaborator for each domain. We also showed that \system{} can act as a general framework for distributed adversarial uDA by incorporating the optimization objectives of various uDA algorithms.

\tabcolsep=0.06cm
\begin{table}[t]
\caption{Mean target accuracy for four uDA methods. \system{} can be used in conjunction with various uDA methods, and improves mean accuracy over the Labeled Source baseline.}
\centering
\scalebox{0.8}{
\begin{tabular}[t]{|c|c|c|c|c||c|c|c|c|} 
\toprule
& \multicolumn{4}{|c||}{RMNIST ($Order_{1}$)} & \multicolumn{4}{c|}{Digits ($Order_{1}$)}  \\
\toprule
&ADDA&GRL&WassDA&CADA&ADDA&GRL&WassDA&CADA\\
\midrule
No Adaptation&34.65&34.65&34.65&34.65&59.59&59.59&59.59&59.59\\
Labeled Source&47.14&47.26&44.39&41.30&64.89&65.51&70.34&65.22\\
Proxy A-distance&93.51&94.91&89.04&78.70&70.09&67.14&72.66&67.0\\
\system{}&\textbf{97.08}&\textbf{97.35}&\textbf{91.15}&\textbf{83.37}&\textbf{73.01}&\textbf{69.80}&\textbf{75.36}&\textbf{70.19}\\
\bottomrule
\end{tabular}
}
\label{tab:gen}
\end{table}

\subsection{Communication Efficiency of \system{}}
\label{results.dils}

To evaluate the communication efficiency of \system{}, we use the amount of data communicated during pairwise adversarial training as a metric. Indeed, if a method exchanges less amount of data between two nodes during distributed training, it will be communication-efficient. Note that in addition to the adversarial training costs, we also incur communication costs during collaborator selection. However, those costs are negligible as compared to adversarial uDA and are not included in our analysis.

Since no previous work has studied the communication costs of adversarial uDA, we choose the most related work FADA \cite{peng2019federated} as our baseline. As discussed in \S\ref{subsec.dils}, FADA was originally designed for multiple sources in a federated learning setup, which is not the case in our setup. For a fair comparison with our single-source setting, we modify FADA by only implementing its Federated Adversarial Alignment component and setting the number of source domains to one. The modified FADA (referred to as FADA$^*$) has the same optimization objectives as single-source DA, but it operates in a distributed setting. Hence, it is fair to compare it with DILS.

\begin{figure}[t]
  \centering
  \includegraphics[width=\linewidth]{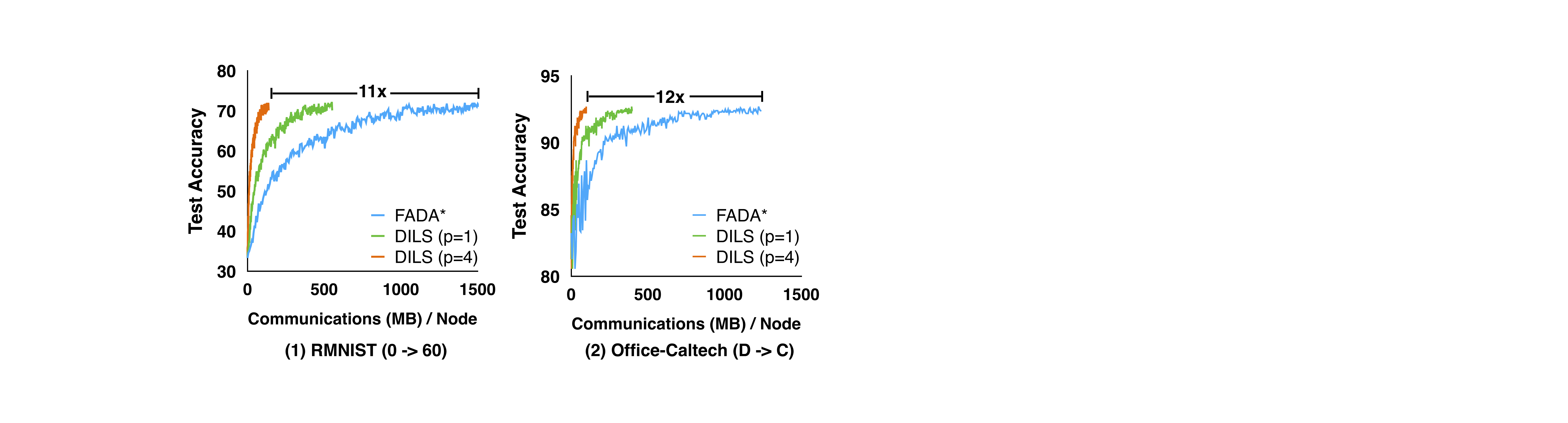}
  \caption{\blue{Test accuracy vs. communication amount per node. DILS (p=4) provides up to $12 \times$ reduction in amount of data exchanged during training.}}
  \label{acc_mb}
\end{figure}

\textbf{Results.} In Figure~\ref{acc_mb}, we plot the test accuracy with the amount of data transmission during pairwise training between two nodes. For DILS, we have two curves with different value of $p$. The results show that with the lazy synchronization strategy ($p=4$), DILS can achieve the same test accuracy as FADA$^*$ or strict synchronized case ($p=1$), but consume much less amount of communication data. On RMNIST and Office dataset, we save up to $12 \times$ traffic compared with FADA$^*$. We also present the similar results for Digits, Mic2Mic and DomainNet in Table~\ref{decuda_acc}. 

\blue{There are two key reasons why DILS can be much more efficient than  FADA$^*$: i) FADA (and FADA$^*$) exchange gradients of the feature extractor during the training, while DILS only exchanges gradients of the discriminator. As the discriminator has significantly fewer parameters than the feature extractor, DILS is inherently more communication-friendly. ii) In DILS ($p=4$), the synchronization between two nodes happens every $p$ steps. Instead, FADA$^*$ requires data exchange after every single step. Therefore, DILS requires much less number of network synchronization. In sum, DILS is able to reduce the communication overhead over FADA$^*$ by exchanging lesser amount of data at a lesser frequency.}

\begin{figure}[t]
\begin{subfigure}{0.5\linewidth}
  \centering
  \includegraphics[scale=0.4]{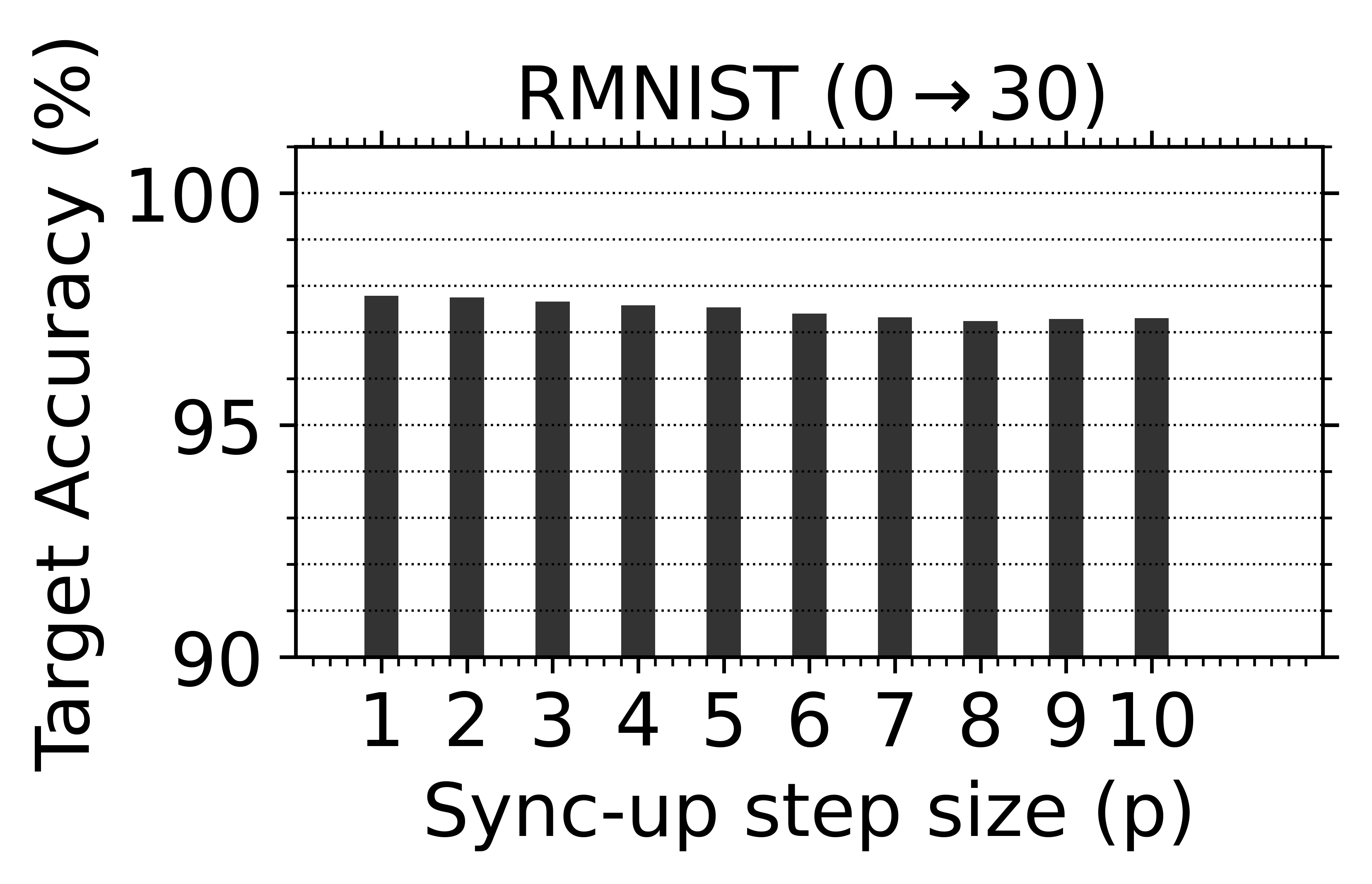}
  \label{fig:syncup-a}
\end{subfigure}%
\hfill
\begin{subfigure}{0.5\linewidth}
  \centering
  \includegraphics[scale=0.4]{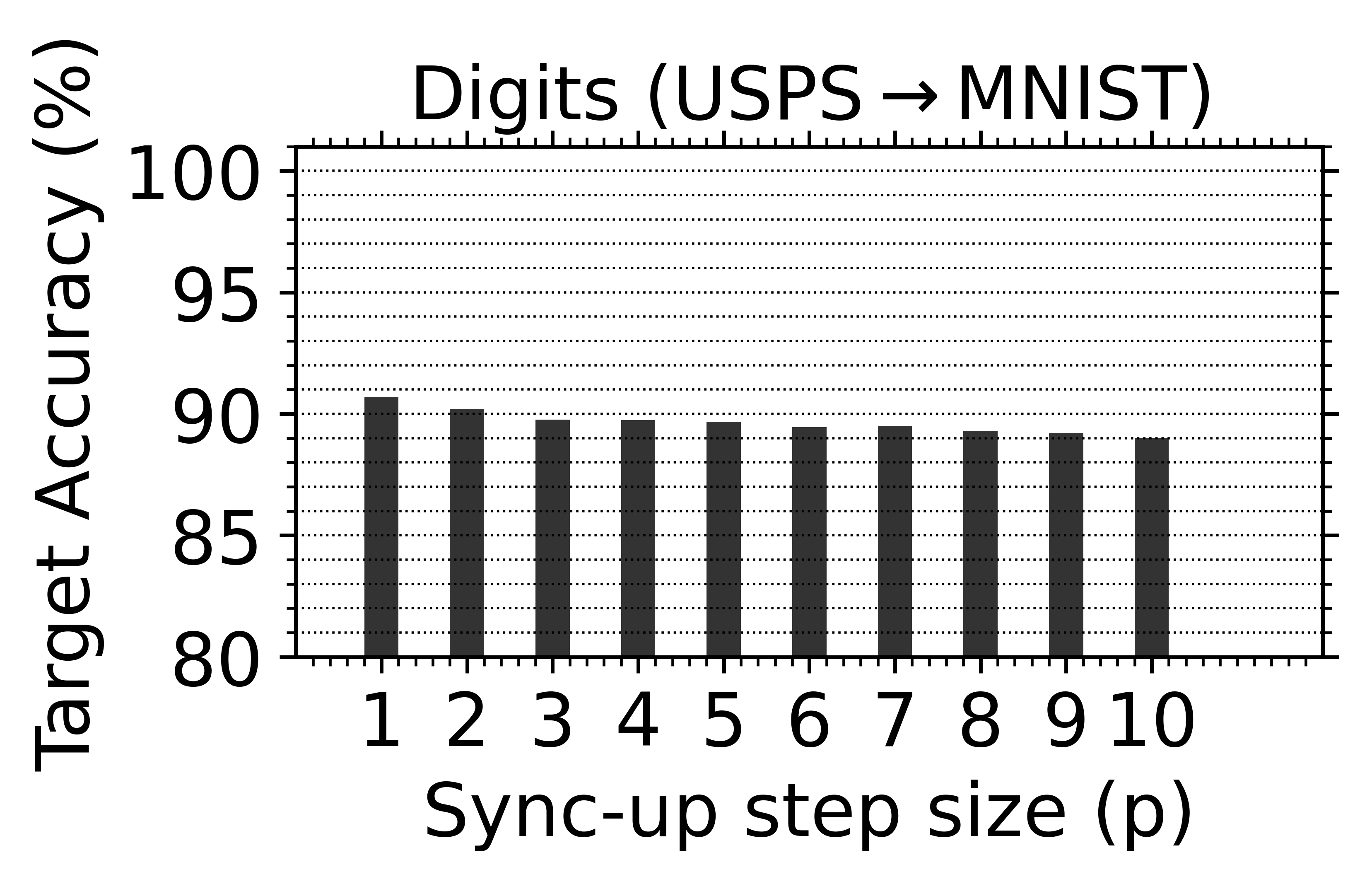}
  \label{fig:syncup-b}
\end{subfigure}
\caption{\blue{Effect of varying the sync-up step $p$ from 1 to 10 on target domain accuracy for two adaptation tasks}}
\label{fig:syncup}
\end{figure}

\parjump{}
\noindent
\textbf{Effect of sync-up step.}
\label{subsec.syncup}
In DILS, the sync-up step size $p$ controls the frequency of gradient exchange between nodes. In Figure~\ref{fig:syncup}, we vary $p$ from 1 and 10 and calculate the adaptation accuracy in the target domain for two adaptation tasks. The results show that DILS is fairly resilient to step-size up to $p=10$. The difference in adaptation accuracy between $p=10$ and $p=1$ is just 0.5\% for RMNIST and 1.7\% for Digits. \blue{This accuracy loss is offset by reduction in data communication and gains in training speed. When $p$ is high, DILS exchanges gradients less frequently, which leads to less communication overhead but slightly worse accuracy. When $p$ is low, the training accuracy is improved at the expense of higher communication cost.}

\blue{Effectively, $p$ could be considered as a tunable parameter to trade-off between target domain accuracy, training time and communication overhead. For applications where it is important to minimize the training time and communication overhead, $p$ could be set to a higher value. Empirically, we find that $p=4$ provides a good tradeoff between accuracy and training speed for all datasets studied in this paper.} 

\begin{table}[t]
\centering
\caption{Communication Amount per node till convergence.}
\small
\begin{tabular}{|c|c|c|c|}
\hline
\textbf{Dataset} & \textbf{Digits} & \textbf{Mic2Mic} & \textbf{DomainNet}\\ \hline
FADA$^*$ & 7.9 GB & 25 GB & 86 GB  \\ 
\hline
$DILS (P=1)$ & 802 MB & 2 GB & 4 GB  \\ \hline
$DILS (P=4)$ & 201 MB & 500 MB & 1 GB  \\ \hline
\end{tabular}
\label{decuda_acc}
\end{table}

\subsection{Analysis and Discussion} 
\label{sec:danda}

We now  analyze some additional properties of \system{}, and provide our thoughts on \system{}'s future research directions.

\parjump{}
\noindent
\textbf{Error accumulation and negative transfer in sequential adaptation.}
\label{subsec.errorprop}
\system{} can be interpreted as a form of sequential adaptation, because new target domains can choose a collaborator from previously adapted target domains using the OCS algorithm. \blue{In sequential adaptation, error accumulation and negative transfer for downstream target domains is a possibility, especially if an unrelated domain appears in the sequence. We analyze OCS from this lens in the following text.} 

\begin{figure}[h]
\centering
\begin{subfigure}{\linewidth}
\centering
  \includegraphics[scale=0.18]{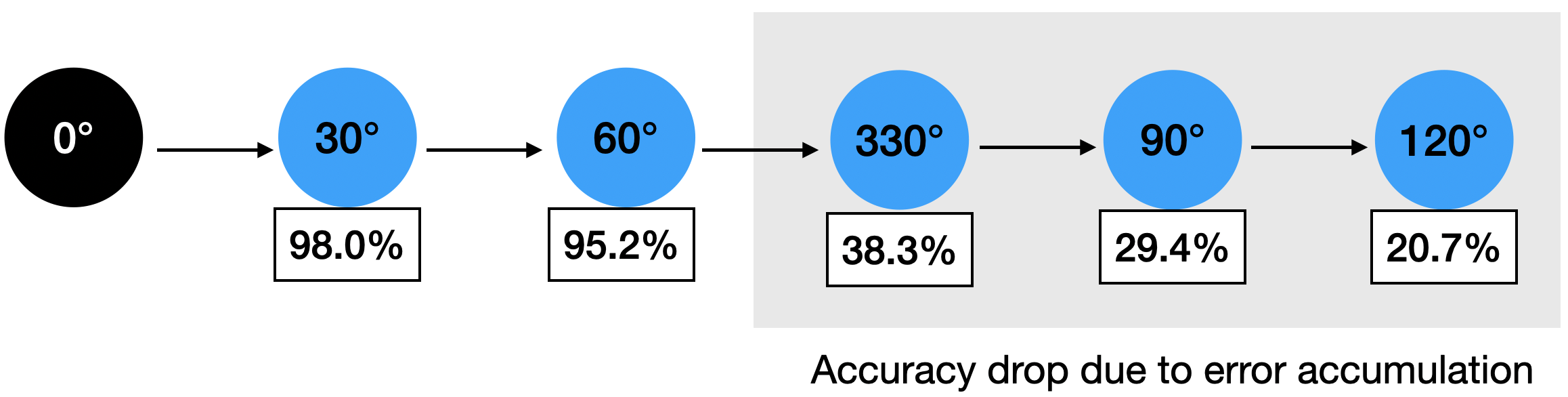}
    \caption{Sequential Adaptation without OCS. Each domain $i$ adapts from the $(i-1)^{th}$ domain.}
  \label{fig:errorprop-a}
\end{subfigure}
\vspace{0.2cm}
\begin{subfigure}{\linewidth}
  \centering
  \includegraphics[scale=0.18]{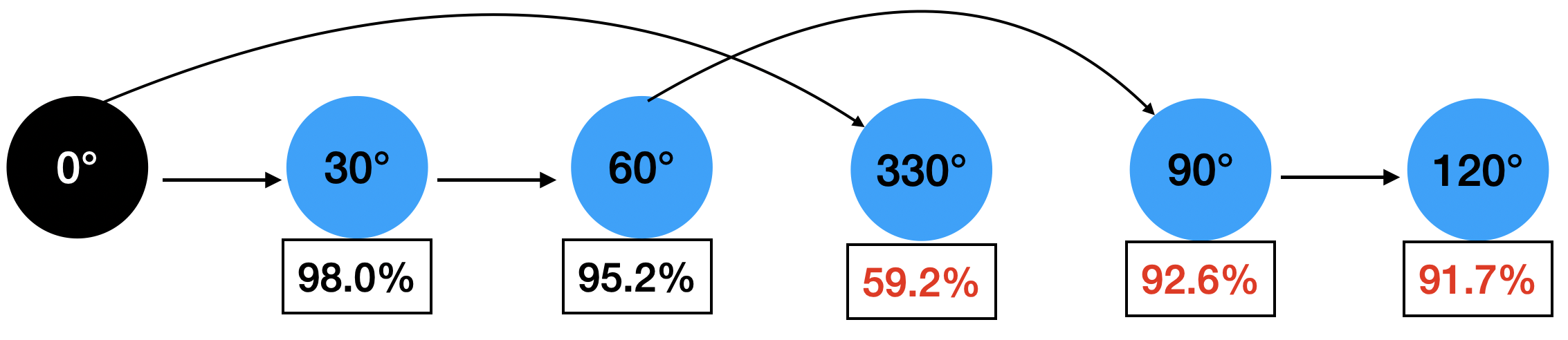}
  \caption{Sequential Adaptation with OCS. Each domain flexibly chooses its optimal collaborator and shows accuracy gains (in red) over the baseline.}
  \label{fig:errorprop-b}
\end{subfigure}
\vspace{-0.5cm}
\caption{OCS prevents negative transfer and error accumulation in sequential adaptation caused by the presence of unrelated domains.}
\vspace{-0.3cm}
\label{fig:errorprop}
\end{figure}

Figure~\ref{fig:errorprop-a} shows a sequence of one source domain (0\textdegree{}) and 5 target domains (30\textdegree{}, 60\textdegree{}, 330\textdegree{}, 90\textdegree{}, 120\textdegree{}) from the RMNIST dataset. \blue{If we simply do a sequential adaptation where each domain $i$ adapts from the $(i-1)^{th}$ domain, we see that 60\textdegree{}$\xrightarrow{}$330\textdegree{} results in high error and poor adaptation accuracy for 330\textdegree. This behavior is due to the high divergence between the two domains. More critically however, we see that this error propagates in all the subsequent adaptation tasks (for 90\textdegree{}, 120\textdegree{}) and causes poor adaptation performance in them as well. In fact, for 120\textdegree{} we also observe negative transfer, as its post-adaptation accuracy (20.7\%) is worse than the pre-adaptation accuracy obtained with source domain model.} 

The ability of OCS to flexibly choose a collaborator for each target domain inherently counters this problem. Firstly, Figure~\ref{fig:errorprop-b} shows that we can choose a better collaborator for 330\textdegree{} using OCS and obtain almost 20\% higher adaptation accuracy than in Figure~\ref{fig:errorprop-a}. More importantly, the subsequent target domains are no longer reliant on 330\textdegree{} as their collaborator and can adapt from any available candidate domain, e.g., 90\textdegree{} adapts from 60\textdegree{} and obtains the best possible adaptation accuracy of 92.6\% in this setup. Similarly, for 120\textdegree{}, we can prevent the negative transfer and achieve 91.7\% accuracy by adapting from 90\textdegree{} (which itself underwent adaptation previously with 60\textdegree{}).

In summary, for a given sequence of domains, OCS enables each target domain to flexibly find its optimal collaborator and achieve the best possible adaptation accuracy. 

\parjump{}
\noindent
\textbf{Future Work}. We showed that \system{} can work with multiple uDA algorithms that follow the adversarial training paradigm introduced in \S\ref{sec:primer}. However, other uDA techniques such as those based on generative algorithms (e.g., \cite{hoffman2018cycada}) and those with no adversarial learning component (e.g., \cite{sun2017correlation}) were out of scope of this paper. Other future works include extending \system{} to scenarios where the label spaces of source and target domains do not overlap~\cite{you2019universal} or when there is label shift between them~\cite{wu2019domain}. 

\blue{\system{}, in its current form, is designed as a sequential adaptation algorithm, that is, it only deals with one target domain at a time. If multiple target domains join together as a batch, \system{} will still process them one by one, and the processing order of the domain could impact their training accuracy. In future work, we will extend \system{} to batch-based processing of target domains. In particular, it will be interesting to explore that given a pool of target domains, how do we sequentially feed them to \system{} such that it maximizes target domain accuracies.}

%% file: chapters/conclusion.tex
In real-world ML applications, datasets are often distributed across thousands of users and devices. Despite the rapid advancements in adversarial uDA research, their extension to distributed settings has surprisingly remained under-explored. We introduced \system{}, an end-to-end framework for distributed adversarial uDA which brings a novel and complementary perspective to uDA research. Through a careful analysis of the literature, we identified the key design requirements for a distributed uDA system and proposed two novel algorithms to increase adaptation accuracy and training efficiency in distributed uDA settings. A comprehensive evaluation with multiple image and speech datasets show that \system{} can increase target accuracy over baselines by as much as 50\% and improve the training efficiency of adversarial uDA by up to $11\times$. Overall, this paper contributes to both domain adaptation and distributed learning literature, by showing for the first time, how domain adaptation and adversarial training can work in a distributed setting.

%% file: chapters/appendix.tex
\section{Reproducibility Details}
\label{sec:implementation}


\subsection{Architectures, Pre-Processing and Hyperparameters} 
\label{subsec.arch}

We now describe the neural architectures used for each dataset along with the pre-processing steps and hyperparameters used for supervised and adversarial learning.  

\textbf{Rotated MNIST:} The MNIST dataset is obtained from the Tensorflow Dataset repository and is rotated by different degrees to generate different domains. The same training and test partitions as in the original dataset are used in our experiments. We employ the LeNet architecture for training the feature extractor. The model was trained for each source domain with a learning rate of $10^{-4}$ using the Adam optimizer and a batch size of 32. 

In the adversarial training process, we used the ADDA loss formulations to perform domain adaptation with a learning rate of $10^{-3}$ for the target extractor and $10^{-2}$ for the discriminators. 

\textbf{Digits:} This task consists of five domains: MNIST, SVHN, USPS, MNIST-Modified and SynthDigits. We used the same train/test split as in the original domain datasets. The images from all domains were normalized between 0 and 1, and resized to 32x32x3 for consistency. The following architecture was used for the feature extractor and it was trained for each source domain with a learning rate of $10^{-5}$ using the Adam optimizer and a batch size of 64. 

\begin{lstlisting}[breaklines]
inputs = tf.keras.Input(shape=(32,32,3), name='img')
x = Conv2D(filters = 64, kernel_size = 5, strides=2)(inputs)
x = BatchNormalization()(x, training=is_training)
x = Dropout(0.1)(x, training=is_training)
x = ReLU()(x)
x = Conv2D(filters = 128, kernel_size = 5, strides=1)(x)
x = BatchNormalization()(x, training=is_training)
x = Dropout(0.3)(x, training=is_training)
x = ReLU()(x)
x = Conv2D(filters = 256, kernel_size = 5, strides=1)(x)
x = BatchNormalization()(x, training=is_training)
x = Dropout(0.5)(x, training=is_training)
x = ReLU()(x)
x = Flatten()(x)
x = Dense(512)(x)
x = BatchNormalization()(x, training=is_training)
x = ReLU()(x)
x = Dropout(0.5)(x, training=is_training)
outputs = Dense(10)(x)
\end{lstlisting}

In the adversarial training process, we used the ADDA losses to perform domain adaptation with a learning rate of $10^{-6}$ for the target extractor and $10^{-4}$ for the discriminator. 

\textbf{Office-Caltech:} We used the pre-trained DeCAF features \cite{decaf} for each domain along with the original train/test splits. The following architecture was used for the feature extractor and it was trained with a learning rate of $10^{-6}$ using the Adam optimizer and a batch size of 32. 

\begin{lstlisting}[breaklines]
Dense(512, activation='linear') 
Dropout(0.7)
Dense(256, activation='linear') 
Dropout(0.7)
Dense(10, activation=None)
\end{lstlisting}

In the adversarial training process, we used the ADDA losses to perform domain adaptation with a learning rate of $10^{-6}$ for the target extractor and $10^{-5}$ for the discriminator.

\textbf{DomainNet:} We used four labeled domains from the DomainNet dataset (Real, Quickdraw, Infograph, Sketch) along with their original train/test splits. A ResNet50-v2 pre-trained on ImageNet was employed as the base model for this task. We froze all but the last four layers of the base model and fine-tuned it for each source domain with a learning rate of 1e-5 using the Adam optimizer and a batch size of 64. 

\begin{lstlisting}[breaklines]
ResNet50V2(include_top=False, input_shape=(224, 224,3), avg='pool'),
Dense(345, activation='softmax')
\end{lstlisting}

In the adversarial training process, we used the ADDA losses to perform domain adaptation with a learning rate of $10^{-6}$ for the target extractor and $10^{-4}$ for the discriminator.

\textbf{Mic2Mic:} Similarly, we followed the same train/test splits as in the original dataset provided by the authors of \cite{mathur2019unsupervised}. The spectrogram tensors were normalized between 0 and 1 during the training and test stages. The following model was trained for each source domain with a learning rate of $10^{-5}$ using the Adam optimizer and a batch size of 64. 

\begin{lstlisting}[breaklines]
Conv2D(filters = 64, kernel_size = (8,20), activation='relu')
MaxPooling2D(pool_size = (2,2)),
Conv2D(filters = 128, kernel_size = (4,10), activation='relu'),
MaxPooling2D(pool_size = (1,4)),
Flatten(),
Dense(256, activation='relu'),
Dense(31)
\end{lstlisting}

In the adversarial training process, we used the ADDA losses to perform domain adaptation with a learning rate of $10^{-3}$ for the target extractor and $10^{-2}$ for the discriminator.

\section{Proofs and Analysis}
\label{sec:proofs}
\input{chapters/proofs}

%% file: chapters/proofs.tex
\subsection{Optimal Collaborator Selection (OCS)} 

\begin{Theorem}
Let $D_1$ and $D_2$ be two domains sharing the same labeling function $l$. Let $\theta_{\mathrm{CE}}$ denote the Lipschitz constant of the cross-entropy loss function in $D_1$, and let $\theta$ be the Lipschitz constant of a hypothesis learned on $D_1$. For any two $\theta$-Lipschitz hypotheses $h, h'$, we can derive the following error bound for the cross-entropy (CE) error $\varepsilon_{\mathrm{CE, D_2}}$ in $D_2$:
\begin{equation}
    \varepsilon_{\mathrm{CE, D_2}}(h, h') \leq  \theta_{\mathrm{CE}} \left( \varepsilon_{\mathrm{L_1, D_1}}(h, h') + 2 \theta W_1(D_1, D_2) \right)
\label{eq:lips}
\end{equation}
where $W_1 (D_1, D_2)$ denote the first Wasserstein distance between the domains $D_1$ and $D_2$, and $\varepsilon_{\mathrm{L_1, D_1}}$ denotes the $L_1$ error in $D_1$.  
\end{Theorem}

\textbf{Proof.} The $L_1$ error between two hypotheses $h, h'$ on a distribution $D$ is given by:
\begin{align}
\label{eq:1_def}
    \varepsilon_{\mathrm{L_1, D}}(h, h') = \mathbb{E}_{{x\sim{}D}} \left[ | h(x) - h'(x) |  \right]
\end{align}

We define softmax cross-entropy on a given distribution $D$ as
\begin{equation}
    \varepsilon_{\mathrm{CE, D}}(h) = \mathbb{E}_{{x\sim{}D}} \left[ \left | \log S_{l(x)} h(x) \right| \right],
\end{equation}
where $S$ is the softmax function $\mathbb{R}^n \longrightarrow \mathbb{R}^n$, $l$ is the labelling function, and $S_{l(x)}$ denotes the projection of $S$ to the $l(x)$-component. 

Then we have, 

\begin{align}
\label{eq:ce_def}
    \varepsilon_{\mathrm{CE, D}}(h, h') &= \mathbb{E}_{{x\sim{}D}} \left[ | \log S_{l(x)} h(x) - \log S_{l(x)} h'(x) |  \right] \nonumber  \\
&= \varepsilon_{\mathrm{L_1, D}}\left( \log S_{l} h, \log S_{l}h' \right)
\end{align}

Further, using the definition of Lipschitz continuity, we have

\begin{equation}
\label{eq:ce_lips}
\left| \log S_{l(x)} h(x) - \log S_{l(y)} h(y) \right| \leq \theta_{\mathrm{CE}} | h(x) - h(y) |, 
\end{equation}
where $\theta_{\mathrm{CE}}$ is the Lipschitz constant of the softmax cross-entropy function.


Next, we follow the triangle inequality proof from \cite[proof of Lemma 1]{shen2018wasserstein} to find that
\begin{align}\label{eq:genineq}
    \varepsilon_{\mathrm{L_1, D_2}}\left( \log S_{l} h, \log S_{l}h' \right) \leq  &\varepsilon_{\mathrm{L_1, D_1}}\left( \log S_{l} h, \log S_{l}h' \right)   \nonumber \\ &+ 2 \theta_{\mathrm{CE}} . \theta W_1(D_1, D_2),
\end{align}
where $\theta$ is a Lipschitz constant for $h$ and $h'$, if the label $l(x)$ were constant. Since $l(x)$ is constant outside of a measure 0 subset where the labels change, and $h$ and $h'$ are Lipschitz, so in particular measurable, Equation \ref{eq:genineq} holds everywhere. 

Then, by substituting from Eq.~\ref{eq:ce_def} and Eq.~\ref{eq:ce_lips} in Eq.~\ref{eq:genineq}, we get Theorem 1:

\begin{align}
 \varepsilon_{\mathrm{CE, D_2}}(h, h') &\leq    \varepsilon_{\mathrm{CE, D_1}}(h, h') + 2 \theta_{\mathrm{CE}} . \theta W_1(D_1, D_2)
  \nonumber \\ 
&\leq \theta_{\mathrm{CE}} (\varepsilon_{\mathrm{L_1, D_1}}\left(h, h'\right) + 2\theta W_1\left(D_1, D_2\right))
\end{align}


\subsection{Convergence of DIscriminator-based Lazy Synchronization (DILS)} 

The network structures of adversarial uDA methods resemble a GAN, where the target encoder $E_T$ and discriminators $D$ ($D_S$ and $D_T$) play a minimax game similar to a GAN's generator and discriminator. $E_S$ and $E_T$ can separately define two probability distributions on the feature representations $E_S(x_s), x_{s} \sim \mathcal X_{S}$ and $E_T(x_t), x_{t} \sim \mathcal X_{T}$, noted as $p_s$ and $p_t$ respectively. Then according to the theoretical analysis in \cite{goodfellow2014generative}, we know that: 

\textbf{Proposition 2.} For a given $E_T$, if $D$ is allowed to reach its optimum, and $p_t$ is updated accordingly to optimize the value function, then $p_t$ converges to $p_s$, which is the optimization goal of $E_T$.

In other words, if we can guarantee that under the training strategy of \textit{Lazy Synchronization}, convergence behaviour of $D_S$ and $D_T$ is similar to the non-distributed case, then $E_T$ should also converge. We have the following theorem.

\textbf{Theorem 1.} In \textit{Lazy Synchronization}, given a fixed target encoder, under certain assumptions, we have the following convergence rate for the discriminators $D_S$ and $D_T$.

\begin{multline}
\frac{1}{T}\sum_{t=1}^{T}\mathbb{E}[\left \| \nabla f(x_t) \right \|_2^2] \leq \frac{1}{1-\mu L} [ \frac{f(x_0)-f(x_{t+1})}{\mu T}+(2p-1)L\mu \sigma ^2 \\ +\frac{\mu L\sigma^2}{2}+\frac{\mu^3L^3\sigma^2(2p-1)}{2}]
\end{multline}

Where $p$ is the sync-up step, $\mu$ is the learning rate. Set $\mu = O(1/\sqrt{T})$. When $p << L\sigma^2$, the impact of stale update will be very small, and thus it can converge with rate $O(1/\sqrt{T})$, which is same as the classic SGD algorithm. 

\textbf{Proof.} As we mentioned in the paper, $D_S$ and $D_T$ are lazily synced, such that their weights are always identical, because they are initialized with the same weights and always apply the same synced gradients at every training step. Therefore, we can consider them as one discriminator $D_{lazy}$ for our convergence analysis.

\emph{Notations.} Throughout the proof, we use following notations and definitions:

\begin{itemize}
    \item $x$ denotes model weights of $D_{lazy}$. 
    \item $f$ denotes the loss function of $D_{lazy}$. 
    \item $\mathcal D_s$ and $\mathcal D_t$ denote the datasets of source and target feature representations (i.e., outputs of the respective feature extractors).
    \item $\xi$ denotes one batch of training instances sampled from $\mathcal D_s \cup \mathcal D_t$.
    \item $f(x, \xi)$ denotes empirical loss of model $x$ on batch $\xi$. 
    \item $\nabla f(\cdot)$ denotes gradients of function $f$.
    \item $\sigma$ denotes the gradient bound.
    \item $\mu$ denotes learning rate.
    \item $T$ denotes the number of training steps.
    \item $p$ denotes the size of sync-up step in our proposed Lazy Synchronization approach. 

\end{itemize}

We can formalize the optimization goal of the discriminator as:
\begin{equation}
\begin{split}
\min_{x \sim \mathbb{R}^N} f(x) ={\mathbb{E}}_{r^{(s)}\sim \mathcal D_s}  \log x(r^{(s)}) + {\mathbb{E}}_{r^{(t)}\sim \mathcal D_t} \log[1 - x(r^{(t)})]
\end{split}
\end{equation}



\emph{Assumption 1.} $f(x)$ is with $L$-Lipschitz gradients:
\begin{equation}
\begin{split}
\left \| \nabla f(x_1)-\nabla f(x_2)\right \|_2^2 \leq L\left \| x_1-x_2 \right \|_2^2 
\end{split}
\end{equation}

Equivalently, we can get:
\begin{equation}
\label{lipschitz}
f(x_2)\leq f(x_1)+\nabla f(x_1)^T(x_2-x_1)+\frac{L}{2}\left \| x_2-x_1 \right \|_2^2
\end{equation}

In training step $t$, we first simplify the updating rule as:
\begin{equation}
\label{update_1}
    x_{t+1} = x_t - \mu \Delta(x_t, \xi_t)
\end{equation}
Combine Inequality \ref{lipschitz} and Equation \ref{update_1}, at time step $t$ we have:
\begin{equation}
\begin{split}
&f(x_{t+1}) \leq f(x_t) - \mu \nabla f(x_t)\Delta(x_t, \xi_t) +\frac{ \mu ^2 L}{2}\left \| \Delta(x_t, \xi_t) \right \|^2_2 \\
&... \\
&f(x_{1}) \leq f(x_0) - \mu \nabla f(x_0)\Delta(x_0, \xi_0) +\frac{ \mu ^2 L}{2}\left \| \Delta(x_0, \xi_0) \right \|^2_2
\end{split}
\end{equation}

Sum all inequalities:
\begin{equation}
    f(x_{t+1}) \leq f(x_0) - \mu \sum_{t=1}^{T}\nabla f(x_t)\Delta(x_t, \xi_t) +\sum_{t=1}^{T} \frac{ \mu ^2 L}{2}\left \| \Delta(x_t, \xi_t) \right \|^2_2
\end{equation}

Take expectation on $\xi_t$ in both sides and re-arrange terms:
\begin{equation}
\label{bound_1}
   \frac{1}{T}\sum_{t=1}^{T} \mathbb{E}[\nabla f(x_t) \Delta(x_t, \xi_t)] \leq \frac{f(x_0)-f(x_{t+1})}{\mu T}+\frac{\mu L}{2T}\sum_{t=1}^{T}\mathbb{E}[\left \| \Delta(x_t, \xi_t) \right \|^2_2]
\end{equation}

In our proposed Lazy Synchronization algorithm, the update term $\Delta(x_t, \xi_t)$ is:

\begin{equation}
\begin{split}
    &\Delta(x_t, \xi_t) = \frac{1}{p}\sum_{n=0}^{p-1}\nabla f(x_{t_p-n}, \xi_{t_p-n}), t\geq p \\
    &t_p = \lfloor t/p \rfloor \times p
\end{split}
\end{equation}

where $t_p$ is the latest sync-up step given a certain time step $t$, and $p$ is the sync-up step of our algorithm. This updating rule formalizes our approach, wherein the gradient applied to the discriminator is the averaged gradient over $p$ steps before the latest sync up step.

We transform $\Delta(x_t, \xi_t)$ as follows:
\begin{equation}
\label{update_2}
\begin{split}
    \Delta(x_t, \xi_t) &= \frac{1}{p}\sum_{n=0}^{p-1}\nabla f(x_{t_p-n}, \xi_{t_p-n})\\
    &= \frac{1}{p}\sum_{n=0}^{p-1}\nabla f(x_{t_p-n},\xi_{t_p-n}) - \frac{1}{p}\sum_{n=0}^{p-1}\nabla f(x_{t},\xi_{t_p-n}) + \\ &\frac{1}{p}\sum_{n=0}^{p-1}\nabla f(x_{t},\xi_{t_p-n}) - \nabla f(x_{t}) + \nabla f(x_{t})\\
    &=\nabla f(x_{t}) + \frac{1}{p}\sum_{n=0}^{p-1}[\nabla f(x_{t_p-n},\xi_{t_p-n}) - \nabla f(x_{t},\xi_{t_p-n})] + \\ &\frac{1}{p}\sum_{n=0}^{p-1}[\nabla f(x_{t},\xi_{t_p-n})-\nabla f(x_{t})]
\end{split}
\end{equation}

Bring Equation \ref{update_2} back to Equation \ref{bound_1} (Let $e(x_t, \xi_t)=\frac{1}{p}\sum_{n=0}^{p-1}[\nabla f(x_{t_p-n},\xi_{t_p-n}) - \nabla f(x_{t},\xi_{t_p-n})] +  \frac{1}{p}\sum_{n=0}^{p-1}[\nabla f(x_{t},\xi_{t_p-n})-\nabla f(x_{t})]$):

\begin{equation}
\label{main_res}
\begin{split}
&\frac{1}{T}\sum_{t=1}^{T} \mathbb{E}[\nabla f(x_t) (\nabla f(x_t) + e(x_t, \xi_t))] \leq \\
&\frac{f(x_0)-f(x_{t+1})}{\mu T}+\frac{\mu L}{2T}\sum_{t=1}^{T}\mathbb{E}[\left \| (\nabla f(x_t) + e(x_t, \xi_t)) \right \|^2_2],\\
&\frac{1}{T}\sum_{t=1}^{T}\mathbb{E}[\left \| \nabla f(x_t) \right \|_2^2]+\frac{1}{T}\sum_{t=1}^{T}\nabla f(x_t)\mathbb{E}[e(x_t, \xi_t)]\leq \\
&\frac{f(x_0)-f(x_{t+1})}{\mu T}+\frac{\mu L}{2T}\sum_{t=1}^{T}\left \| \nabla f(x_t) \right \|_2^2+\frac{\mu L}{2T}\sum_{t=1}^{T}\mathbb{E}[\left \| e(x_t, \xi_t)  \right \|_2^2] ,\\
&\frac{1}{T}\sum_{t=1}^{T}\mathbb{E}[\left \| \nabla f(x_t) \right \|_2^2]\leq \frac{2}{2-\mu L} \biggr( \frac{f(x_0)-f(x_{t+1})}{\mu T} - \\
&\frac{1}{T}\sum_{t=1}^{T}\nabla f(x_t)\mathbb{E}[e(x_t, \xi_t)] + \frac{\mu L}{2T}\sum_{t=1}^{T}\mathbb{E}[\left \| e(x_t, \xi_t)  \right \|_2^2] \biggr).
\end{split}
\end{equation}

Then we analyze $\mathbb{E}[e(x_t, \xi_t)]$ and $\mathbb{E}[\left \| e(x_t, \xi_t))  \right \|_2^2]$:

\begin{equation}
\begin{split}
    e(x_t, \xi_t)=\underbrace{\frac{1}{p}\sum_{n=0}^{p-1}[\nabla f(x_{t_p-n},\xi_{t_p-n}) - \nabla f(x_{t},\xi_{t_p-n})]}_{e_1} + \\ \underbrace{\frac{1}{p}\sum_{n=0}^{p-1}[\nabla f(x_{t},\xi_{t_p-n})-\nabla f(x_{t})]}_{e_2}
\end{split}
\end{equation}

For $e_1$:
\begin{equation}
\label{e1}
\begin{split}
    e_1(x_t, \xi_t)=&\frac{1}{p}\sum_{n=0}^{p-1}[\nabla f(x_{t_p-n},\xi_{t_p-n}) - \nabla f(x_{t},\xi_{t_p-n})]\\
    =&\frac{1}{p}\sum_{n=0}^{p-1}[\nabla f(x_{t_p-n},\xi_{t_p-n})- \nabla f(x_{t_p-n+1},\xi_{t_p-n}) + \\
    &\nabla f(x_{t_p-n+1},\xi_{t_p-n}) - ... -\nabla f(x_{t_p},\xi_{t_p-n}) \\
    &+ \nabla f(x_{t_p},\xi_{t_p-n})-...+ \nabla f(x_{t-1},\xi_{t_p-n}) \\
    &- \nabla f(x_{t},\xi_{t_p-n})]\\
    =&\frac{1}{p}\sum_{n=0}^{p-1}\biggr (\sum_{i=0}^{n-1}[\nabla f(x_{t_p-(i+1)},\xi_{t_p-n})-\nabla f(x_{t_p-i},\xi_{t_p-n})] +\\
    &\sum_{j=0}^{t-t_p-1}[\nabla f(x_{t_p+j},\xi_{t_p-n})-\nabla f(x_{t_p+j+1},\xi_{t_p-n})] \biggr)
\end{split}
\end{equation}

According to the algorithm, we know that:
\begin{equation}
\begin{split}
x_{t_p-i} &= x_{t_p-(i+1)} - \mu G_{(\lfloor t/p \rfloor -1 )\times p}\\
x_{t_p+j+1} &= x_{t_p+j} - \mu G_{\lfloor t/p \rfloor \times p}
\end{split}
\end{equation}
where $G_{(\lfloor t/p \rfloor -1 )\times p}$ is the second last synced gradient, $G_{\lfloor t/p \rfloor \times p}$ is the latest synced gradient.

Apply L-Lipschitz gradient:
\begin{equation}
\begin{split}
\left \| \nabla f(x_{t_p-(i+1)},\xi_{t_p-n})-\nabla f(x_{t_p-i},\xi_{t_p-n}) \right \|&\leq L\mu\left \|  G_{(\lfloor t/p \rfloor -1 )\times p}\right \| \\
\left \| \nabla f(x_{t_p+j},\xi_{t_p-n})-\nabla f(x_{t_p+j+1},\xi_{t_p-n}) \right \|&\leq L\mu\left \|  G_{\lfloor t/p \rfloor\times p}\right \|
\end{split}
\end{equation}
\vspace{0.3cm}

\emph{Assumption 2.} Bounded Gradient. We assume the stochastic gradients are uniformly bounded: (see \cite{shalev2011pegasos, nemirovski2009robust, hazan2014beyond})
\begin{equation}
    \mathbb{E}[\left \| \nabla f(x_t,\xi_t) \right \|_2^2]\leq \sigma^2
\end{equation}

Therefore, we have:
\begin{equation}
\label{bounded_variance}
\begin{split}
\left \| \nabla f(x_{t_p-(i+1)},\xi_{t_p-n})-\nabla f(x_{t_p-i},\xi_{t_p-n}) \right \|&\leq L\mu\ \sigma \\
\left \| \nabla f(x_{t_p+j},\xi_{t_p-n})-\nabla f(x_{t_p+j+1},\xi_{t_p-n}) \right \|&\leq L\mu \sigma
\end{split}
\end{equation}

Bring Inequality \ref{bounded_variance} back to Equality \ref{e1}:

\begin{equation}
\begin{split}
    e_1(x_t, \xi_t)&\leq \frac{1}{p}\sum_{n=0}^{p-1}(n+t-t_p)L\mu \sigma \\
    &\leq (2p-1)L\mu \sigma
\end{split}
\end{equation}

since it is easy to see $t-t_p\leq p$.

In summary of $e_1$, we get:
\begin{equation}
\label{e1_final}
\begin{split}
   \mathbb{E}[\nabla f(x_t)e_1(x_t, \xi_t)]&\leq \mathbb{E}[\left \|\nabla f(x_t)  \right \|] \times \left \|e_1(x_t, \xi_t)  \right \|]\leq (2p-1)L\mu \sigma^2 \\
    \mathbb{E}[\left \| e_1(x_t, \xi_t))  \right \|_2^2]&\leq (2p-1)L^2\mu^2 \sigma^2
\end{split}
\end{equation}

For $e_2= \frac{1}{p}\sum_{n=0}^{p-1}[\nabla f(x_{t},\xi_{t_p-n})-\nabla f(x_{t})]$:

Since batch $\xi$ is an unbiased sampled batch, it is obvious that for a certain $n$:
\begin{equation}
\begin{split}
\mathbb{E}[\nabla f(x_{t},\xi_{t_p-n})-\nabla f(x_{t})] &= 0 
\end{split}
\end{equation}

Also,
\begin{equation}
\begin{split}
\mathbb{E}[\left \| \nabla f(x_{t},\xi_{t_p-n})-\nabla f(x_{t}) \right \|_2^2] &\leq \mathbb{E}[\left \|\nabla f(x_{t})  \right \|_2^2] + \mathbb{E}[\left \| \nabla f(x_{t},\xi_{t_p-n}) \right \|_2^2]  \\
&\leq \mathbb{E}[\left \|\nabla f(x_{t})  \right \|_2^2] + \sigma^2
\end{split}
\end{equation}

Therefore,
\begin{equation}
\label{e2}
\begin{split}
\mathbb{E}[e_2(x_{t},\xi_{t})]&= 0 \\
\mathbb{E}[\left \| e_2(x_t, \xi_t))  \right \|_2^2]&\leq \mathbb{E}[\left \|\nabla f(x_{t})  \right \|_2^2] + \sigma^2
\end{split}
\end{equation}

Combine \ref{e1_final} and \ref{e2} with \ref{main_res}:
\begin{multline}
\frac{1}{T}\sum_{t=1}^{T}\mathbb{E}[\left \| \nabla f(x_t) \right \|_2^2] \leq \frac{1}{1-\mu L} [ \frac{f(x_0)-f(x_{t+1})}{\mu T}+ \\(2p-1)L\mu \sigma ^2+\frac{\mu L\sigma^2}{2}+\frac{\mu^3L^3\sigma^2(2p-1)}{2}]
\end{multline}

Therefore, the expected $\left \| \nabla f(x_t) \right \|_2^2$ converge to 0 with rate $O(1/\sqrt{T})$ if $\mu = 1/\sqrt{T}$. When the function $f(x)$ is strongly convex, this is the global minimum, otherwise it can be a local minimum, or a stationary point.